\documentclass{article}

\usepackage{iclr2027_conference,times}

\usepackage{amsmath,amsfonts,bm}

\def\eqref#1{equation~\ref{#1}}

\def\1{\bm{1}}

\DeclareMathAlphabet{\mathsfit}{\encodingdefault}{\sfdefault}{m}{sl}
\SetMathAlphabet{\mathsfit}{bold}{\encodingdefault}{\sfdefault}{bx}{n}

\usepackage{graphicx}
\usepackage{wrapfig}
\usepackage{subcaption}

\usepackage{amsmath}
\usepackage{amssymb}
\usepackage{amsthm}

\usepackage{booktabs}
\usepackage{siunitx}
\usepackage{threeparttable}
\usepackage[table]{xcolor}

\usepackage{algorithm}
\usepackage{algpseudocode}

\usepackage{csquotes}
\usepackage{pifont}
\usepackage{url}

\usepackage{hyperref}
\usepackage[capitalise]{cleveref}

\newtheorem{theorem}{Theorem}[section]

\newtheorem{proposition}[theorem]{Proposition}
\newcommand{\hifmrestatedname}{}
\newtheorem*{hifmrepeatedproposition}{\hifmrestatedname}
\theoremstyle{remark}
\newtheorem{remark}[theorem]{Remark}

\theoremstyle{plain}
\crefname{proposition}{Proposition}{Propositions}
\Crefname{proposition}{Proposition}{Propositions}
\crefname{remark}{Remark}{Remarks}
\Crefname{remark}{Remark}{Remarks}

\definecolor{beliz}{RGB}{41, 128, 185}
\definecolor{nephritis}{RGB}{39, 174, 96}
\definecolor{alazarin}{RGB}{231, 76, 60}

\providecommand{\hifmd}{\textsc{HI-FM-D}}
\providecommand{\hifmt}{\textsc{HI-FM-T}}
\providecommand{\hifme}{\textsc{HI-FM-E}}

\title{Hessian-Informed Flow Matching}

\author{%
\textbf{Christopher Iliffe Sprague}\textsuperscript{*, 1}\qquad
\textbf{Mert Can Kurucu}\textsuperscript{*, 2, 3}\qquad
\textbf{Riccardo Tedoldi}\textsuperscript{2,4}\\[0.35em]
\textbf{Alessandro Tibo}\textsuperscript{4}\qquad
\textbf{Jon Paul Janet}\textsuperscript{4}\qquad
\textbf{Hossein Azizpour}\textsuperscript{2, 5}\\[0.6em]
{\normalfont
\textsuperscript{1}Orbital Industries,\quad
\textsuperscript{2}KTH Royal Institute of Technology}\\
{\normalfont
\textsuperscript{3}Istanbul Technical University,\quad
\textsuperscript{4}AstraZeneca}\\
{\normalfont
\textsuperscript{5}Science for Life Laboratory}
}

\iclrfinalcopy

\begin{document}

\maketitle

\begingroup
\renewcommand{\thefootnote}{*}
\footnotetext{Equal contribution. Code available at
\url{https://github.com/mertcankurucu/hi-fm}.}
\endgroup

\fancyhead[L]{Preprint}

\begin{abstract}
Data from physical systems often concentrate near stable states
corresponding to minima of an energy function. Around these states, the curvature of the energy landscape
shapes the local covariance structure of the data, which is
often highly anisotropic: small perturbations around a data point
raise its energy more sharply in some directions than in others. However, standard flow
matching commonly connects noise to target data through isotropic
conditional paths that do not explicitly follow this physical
structure. We introduce Hessian-Informed Flow Matching (HI-FM), which
uses the energy Hessian at each data sample to encode this local
curvature in the conditional probability paths used for training.
By locally linearizing energy-driven dynamics, we derive anisotropic
paths that move samples toward their targets at different rates in different directions, without changing the standard flow matching
objective. We evaluate HI-FM through molecular generation on
Hessian-QM9 and GEOM-Drugs, using SemlaFlow as a shared backbone. With Hessians calculated offline only for the training data,
across both datasets, HI-FM improves the physical and geometric
quality of generated molecules while maintaining comparable chemical
validity. Under a matched optimizer-update budget, HI-FM
trained with only $10\%$ of Hessian-QM9 achieves lower mean
energy, strain, and bond-length $W_1$ than the full-data SemlaFlow
baseline. The results support local energy curvature as a useful
inductive bias for flow matching.
\end{abstract}

\section{Introduction}

Generative modeling is a central problem in machine learning, where the primary goal is to learn a model that can generate samples from a complex target distribution.
In recent years, flow-based generative models, such as diffusion \citep{Song2020ScoreBasedGM} and flow matching (FM)~\citep{Lipman2022FlowMF} have come to the forefront of deep generative modeling, with success ranging from image generation \citep{Rombach2021HighResolutionIS} to biology applications \citep{abramson2024accurate} to robotics \citep{chi2024diffusionpolicy}.

Among these models, flow matching learns to transport a simple source distribution to data by regressing a vector field on samples from conditional probability paths. The standard linear conditional path contracts an isotropic Gaussian source equally in every direction as it moves toward a data sample. This is not desirable as in physical systems, one is often interested in identifying equilibrium states that likely exhibit anisotropic dynamics.
For instance, in biology, different conformations in protein folding \citep{abramson2024accurate} and drug binding \citep{Corso2022DiffDockDS} represent minima of a potential energy function. Similarly, in robotics, stable formations of robotic swarms \citep{sun2017rigid} or stable grasps of robotic manipulators \citep{jiang2021hand} often minimize an energy function, for example a Lyapunov function \citep{la1966invariance}.
More generally, in statistical physics, it is of interest to identify energy-minimizing states that, in turn, are likelihood-maximizing states.

The states of these systems are often considered to evolve according to the gradient descent of some potential function.
These dynamics eventually converge to a stationary distribution (about one of the energy's minima), whose covariance is determined by the local curvature of the energy landscape, and is often \textit{anisotropic}.
Despite that both flow matching and the physical systems whose equilibrium distribution they often seek to sample, are based on dynamical systems, discrepancies between them remain: most flow models only employ \textit{isotropic} probability paths, ignoring potentially important aspects of the underlying physics. In this work, we address this limitation.

Properly accounting for the anisotropy inherent to a given system is important for the following reasons.
It captures local energy curvature, which may help models generalize to unseen regions of the ambient space.
It captures different dynamical timescales, with dynamics converging more slowly in some directions and faster in others.
It also reflects energy invariances and the equivariance of gradient dynamics: at a non-collinear equilibrium configuration, the Hessian of an energy function invariant to 3D translations and rotations has at least six zero eigenvalues corresponding to infinitesimal rigid-body motions.

\begin{figure*}[t]
    \centering
    \includegraphics[width=\textwidth]{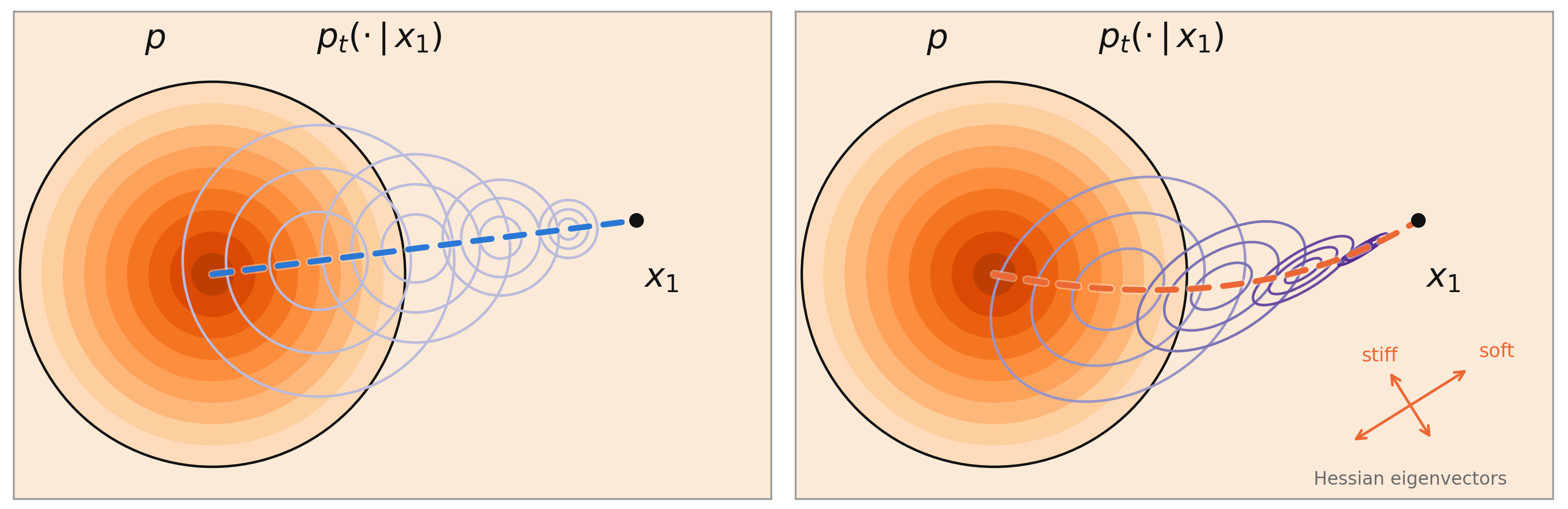}
    \caption{
        \textbf{Linear and Hessian-informed conditional paths.}
        Both paths connect the same isotropic source distribution $p$
        to a data sample $\mathbf{x}_1$.
        \textbf{Left (linear path):} The conditional distribution narrows
        equally in all directions, while its mean follows a straight line
        toward $\mathbf{x}_1$.
        \textbf{Right (HI-FM):} The Hessian of the energy at $\mathbf{x}_1$
        determines how the conditional distribution changes along different
        directions. Its contours align with the Hessian eigenvectors and
        become narrower along stiff (high-curvature) directions than along
        soft (low-curvature) directions.
    }
    \label{fig:conditional_paths}
\end{figure*}

In this work, we construct anisotropic conditional paths that incorporate
local energy landscape information through the energy Hessian at each
training sample (\cref{fig:conditional_paths}). We base this construction
on the well-known linearization theorem from dynamical systems
\citep{hartman1960lemma}: we evolve the Gaussian source under gradient
dynamics linearized at the data sample and map the resulting relaxation
onto a finite flow-time interval. The resulting Hessian-Informed Flow
Matching (HI-FM) paths contract stiff directions faster than soft ones
while keeping the source and data endpoints unchanged. HI-FM changes only
the conditional paths and their velocity targets: the FM objective,
network, and sampler are unchanged, Hessians are precomputed for the
training data, and generation requires no energy or Hessian evaluations.

Our experiments with SemlaFlow~\citep{irwin2024semlaflow} on Hessian-QM9 and GEOM-Drugs show
that HI-FM improves molecular generation quality, data efficiency,
and optimization efficiency. Across both datasets, HI-FM reduces mean energy, strain, and
bond-length and angle $W_1$, while maintaining competitive validity. With full training data, mean strain is reduced
by $14.4$-$17.3\%$ on Hessian-QM9 and $9.6$-$15.0\%$ on
GEOM-Drugs relative to the corresponding SemlaFlow baselines,
demonstrating gains on both small and larger, more diverse molecules.
HI-FM is also more data-efficient: on Hessian-QM9, 
trained with only $10\%$ of the data HI-FM achieves lower mean energy,
strain, and bond-length $W_1$ than full-data SemlaFlow under the
same optimizer-update budget. Training curves show that HI-FM reaches the baseline's energy and
strain earlier on both datasets, with approximately $23$-$31\%$
fewer optimizer updates in the $10\%$ Hessian-QM9 setting.
Our timing benchmark shows that using precomputed Hessians adds
$18.9\%$ time per optimizer update for direct HI-FM.

\section{Related Work}

Flow-based deep generative models have gained significant attention for their ability to model complex distributions in various domains. Notably, diffusion models \citep{Song2020ScoreBasedGM} have achieved state-of-the-art results in image generation tasks \citep{dhariwal2021diffusion} and have been extended to applications such as structural biology \citep{Corso2022DiffDockDS, Yim2023SE3DM, Ketata2023DiffDockPPRP} and video generation \citep{Ho2022VideoDM, Blattmann2023AlignYL, Esser2023StructureAC}, incorporating latent representations \citep{Vahdat2021ScorebasedGM, Blattmann2023AlignYL} and geometric priors \citep{Bortoli2022RiemannianSG, Dockhorn2021ScoreBasedGM}.

Flow matching models \citep{Lipman2022FlowMF} have been proposed as an alternative to diffusion models, offering faster training and sampling while maintaining competitive performance. Relying on continuous normalizing flows \citep{Chen2018NeuralOD}, FM models generalize diffusion models, as demonstrated by the existence of the probability flow ODE that induces the same marginal probability density function as the Stochastic Differential Equation (SDE) of diffusion models \citep{Song2020ScoreBasedGM}. 
FM models have found applications in structural biology and molecular generation~\citep{Yim2023FastPB, Bose2023SE3StochasticFM, tedoldi2026flexiflow}, media~\citep{Le2023VoiceboxTM, Liu2023I2SBIS}, and have been extended in various ways~\citep{Tong2023ConditionalFM, Pooladian2023MultisampleFM, Shaul2023OnKO, Chen2023RiemannianFM, Klein2023EquivariantFM}.

In modeling physically stable states, such as molecular
conformations, several works have leveraged the relationship
between equilibrium density scores and energy gradients under Boltzmann models. For instance, \citet{zaidi2022pre} highlighted the equivalence between denoising score matching \citep{Vincent2011ACB} and force-field learning. This concept was extended in \citet{Feng2023MayTF} to incorporate off-equilibrium data and neural network gradient fields. Other works \citep{Shi2021LearningGF, Luo2021PredictingMC} have learned neural network gradient fields to model pseudo-force fields, subsequently using them to sample energy-minimizing molecular conformations via annealed Langevin dynamics \citep{Song2019GenerativeMB}. More recently, \citet{Zhou2025EnergyBasedFM} adopt an energy-based perspective for flow matching in 3D molecular structure generation, but do not construct conditional paths from target energy Hessians.

While several of these approaches leverage energy gradients or learned gradient fields, most flow-based generative modeling approaches assume isotropic covariance structures and may not fully capture the anisotropic characteristics arising from the curvature of the energy landscape. Only a few approaches have considered non-isotropic distributions \citep{yu2024constructing, singhal2024s}, but these do not deal with Hessians. \citet{dehmamy2024hessian} have considered gradient flows defined by Hessians, but not for flow-based generative models. The energy Hessian provides a direct way to construct anisotropic
conditional paths from local energy curvature.
In contrast, HI-FM uses the energy Hessian to construct anisotropic conditional paths for flow matching, with the construction grounded in local dynamical-systems theory.

\section{Preliminaries}
\label{sec:preliminaries}

Let $q$ denote the data distribution on $\mathbb{R}^d$ and $p_0$
a simple prior distribution (e.g., Gaussian). FM~\citep{Lipman2022FlowMF} learns a time-dependent vector field (VF)
$\mathbf{v}_\theta$ to transport $p_0$ toward $q$. Samples are
generated by integrating
$\dot{\mathbf{x}}_t=\mathbf{v}_\theta(\mathbf{x}_t,t)$ from
$\mathbf{x}_0\sim p_0$ over $t\in[0,1]$, with the data
endpoint approached as $t\to1$.

\paragraph{Matching conditional flows.}
For each data sample $\mathbf{x}_1\sim q$, consider a conditional
probability path $p_t(\mathbf{x}\mid\mathbf{x}_1)$ starting from
$p_0$ and concentrating at $\mathbf{x}_1$ as $t\to1$, with a known
transporting VF $\mathbf{u}_t(\mathbf{x}\mid\mathbf{x}_1)$.
Conditional flow matching trains the model using
\begin{equation}
    \mathcal{L}_{\mathrm{CFM}}(\theta)
    =\mathbb{E}_{t,\mathbf{x}_1,\mathbf{x}_t}
    \left[
        \left\|\mathbf{v}_\theta(\mathbf{x}_t,t)
        -\mathbf{u}_t(\mathbf{x}_t\mid\mathbf{x}_1)\right\|_2^2
    \right],
    \label{eq:fm}
\end{equation}
where $t\sim\mathcal{U}(0,1)$, $\mathbf{x}_1\sim q$, and
$\mathbf{x}_t\sim p_t(\cdot\mid\mathbf{x}_1)$.
Under standard regularity conditions, this objective has the same
parameter gradients as regression onto the VF transporting
the marginal path obtained by averaging the conditional paths over
$q$~\citep{Lipman2022FlowMF}.

\paragraph{Linear conditional paths.}
For independent samples $\mathbf{x}_0\sim p_0$ and
$\mathbf{x}_1\sim q$, the linear conditional path
with zero terminal noise and its velocity target are
\begin{equation}
    \mathbf{x}^{\mathrm{lin}}_t
    =(1-t)\mathbf{x}_0+t\mathbf{x}_1,
    \qquad
    \partial_t\mathbf{x}^{\mathrm{lin}}_t
    =\mathbf{x}_1-\mathbf{x}_0.
    \label{eq:linear_conditional_path}
\end{equation}
For $p_0=\mathcal{N}(\mathbf{0},\mathbf{I})$, this induces
\begin{equation}
    p_t^{\mathrm{lin}}(\mathbf{x}\mid\mathbf{x}_1)
    =\mathcal{N}\!\left(t\mathbf{x}_1,(1-t)^2\mathbf{I}\right),
    \qquad t<1.
    \label{eq:linear_conditional_gaussian}
\end{equation}
Thus, the conditional mean follows a straight line toward the target data,
while the conditional spread decreases equally in every direction.

\section{Hessian-Informed Flow Matching}
\label{sec:hifm}

Near a stable molecular conformation, small deformations of equal magnitude can have different energy costs, whereas the linear conditional path uses the same contraction schedule in every direction, ignoring potentially important aspects of the underlying physics. Our aim is to make the FM conditional dynamics reflect this local energy structure; we then test empirically whether doing so improves the energetic and geometric quality of generated molecules (\cref{sec:experiments}). We propose using energy Hessians, at the equilibria, to incorporate the local energy
structure into the conditional probability paths used for FM. For each data sample $\mathbf{x}_1$, we construct
these paths from a local linear approximation of energy-driven
dynamics at the equilibria. Evolving a Gaussian source under these dynamics gives
conditional distributions whose mean and covariance reflect the local energy curvature at $\mathbf{x}_1\sim q$. The resulting paths and their
associated conditional VFs provide the training samples
with updated velocity targets for \cref{eq:fm} as we will see next.

\subsection{Hessian-Informed Conditional Paths}
\label{sec:hessian_paths}

\paragraph{Local dynamics from the Hessian.}
We consider data near minima of a smooth energy function $V$ and
assume access to its Hessian at each data sample $\mathbf{x}_1$.
Its eigendecomposition is
\begin{equation}
    \mathbf{H}_1
    =\nabla_{\mathbf{x}}^2V(\mathbf{x}_1)
    =\mathbf{P}\operatorname{diag}(\boldsymbol{\lambda})
      \mathbf{P}^{\top},
    \label{eq:target_hessian}
\end{equation}
where $\boldsymbol{\lambda}=(\lambda_1,\ldots,\lambda_d)^{\top}$
contains the eigenvalues of $\mathbf{H}_1$, and the columns
of $\mathbf{P}$ are the corresponding orthonormal eigenvectors. Each eigenvector defines a local deformation
direction, and its eigenvalue measures the energy curvature along
that direction. For small displacements of equal magnitude at a
minimum, a larger positive eigenvalue corresponds to a larger
energy increase. These high-curvature directions are \emph{stiff},
whereas directions with smaller positive eigenvalues are \emph{soft}.
For example, in molecular generation tasks, a deformation dominated by bond stretching can have
a much larger positive Hessian eigenvalue than one dominated by
torsion (see \cref{fig:co_stretch_alignment} in the appendix for a C=O-stretching example). These curvature differences also determine how
deformations relax under local gradient dynamics. Treating
$\mathbf{x}_1$ as an equilibrium, the local approximation
$-\nabla_{\mathbf{x}}V(\mathbf{x})
\approx-\mathbf{H}_1(\mathbf{x}-\mathbf{x}_1)$
gives exponential decay at rate $\lambda_i$ for displacement
along each eigenvector with $\lambda_i>0$.
Stiff deformations therefore relax faster than soft ones.
The linear conditional path in \cref{eq:linear_conditional_path},
by contrast, reduces displacements in every direction by the
same factor $1-t$, regardless of energy curvature.

\paragraph{How the conditional distribution changes.}
In conditional flow matching, the conditional probability path
determines the intermediate distributions used for training.
To make these distributions reflect local energy curvature,
we replace the linear interpolation in
\cref{eq:linear_conditional_path} with evolution under the
Hessian-based dynamics at the data sample. For a fixed data sample $\mathbf{x}_1$,
these linear dynamics preserve the Gaussian form of the source,
giving the conditional probability path
\begin{equation}
    p_t^{\mathrm H}(\mathbf{x}\mid\mathbf{x}_1)
    =\mathcal{N}\!\left(
        \boldsymbol{\mu}_{\mathrm H}(t),
        \boldsymbol{\Sigma}_{\mathrm H}(t)
      \right),
    \label{eq:hifm_conditional_gaussian}
\end{equation}
where $\boldsymbol{\mu}_{\mathrm H}(t)$ and
$\boldsymbol{\Sigma}_{\mathrm H}(t)$ are the conditional mean
and covariance determined by the Hessian-based dynamics (see \cref{sec:interpolant} for the derivation).
The mean approaches the target data sample at different rates along the
Hessian eigenvectors, rather than following the same schedule
in every direction. Without added diffusion ($\mathbf{B}_1=\mathbf{0}$ in \cref{sec:interpolant}), the covariance reflects these
directional rates as well:
starting from an isotropic source, the distribution becomes
narrower along stiff directions than along soft ones, producing
Hessian-aligned ellipsoidal contours rather than the spherical
contours of the linear path (\cref{fig:conditional_paths}).
The point-target construction used in our experiments preserves
the prior distribution and converges to the same target
$\mathbf{x}_1$ as $t\to1$; the Hessian changes the intermediate
distributions.

\subsection{Hessian-Informed Path Construction}\par We construct the HI-FM path by linearizing energy-driven dynamics at each target sample and mapping the resulting relaxation onto finite flow time. In the Hessian eigenbasis, the local approximation $-\nabla_{\mathbf{x}}V(\mathbf{x})\approx-\mathbf{H}_1(\mathbf{x}-\mathbf{x}_1)$ decouples the dynamics, so displacements along eigenvector $i$ decay at a rate determined by its curvature. For the path used in our experiments, we replace the raw eigenvalues $\lambda_i$ by positive processed rates $\alpha_i$ while retaining the Hessian eigenvectors. \par Mapping relaxation time $s$ to flow time using $t=1-\exp(-\kappa\alpha_{\min}s)$ then gives the directional exponents \begin{equation} r_i=\frac{\alpha_i}{\kappa\alpha_{\min}},\qquad \alpha_{\min}=\min_i\alpha_i, \label{eq:hifm_rates_main} \end{equation} where $\kappa>0$ controls the time scaling; the processing of $\alpha_i$ is described below. The full stochastic-system, Gaussian-moment, probability-flow, and time-reparameterization derivation is given in \cref{app:hifm_derivation}.\par For the zero-diffusion point-target construction used in our experiments, the resulting conditional sampling map is \begin{equation} \mathbf{x}^{\mathrm H}_t=\mathbf{x}_1+\mathbf{P}\operatorname{diag}\!\left((1-t)^{r_i}\right)\mathbf{P}^{\top}(\mathbf{x}_0-\mathbf{x}_1),\qquad \mathbf{x}_0\sim p_0. \label{eq:hifm_sample_path_main} \end{equation} Differentiating this map gives the conditional velocity target in \cref{eq:fm}. At $t=0$ the map is the identity and therefore preserves $p_0$; because every $r_i>0$, it converges to $\mathbf{x}_1$ as $t\to1$. For $\kappa=1$, equal rates recover the linear conditional path in \cref{eq:linear_conditional_path}, whereas unequal rates contract stiff directions faster than soft ones.\par Molecular Hessians contain null modes associated with rigid translations and rotations. We rescale eigenvalues above a positive threshold to control the spread of relaxation rates and use \emph{hyperbolization} (\cref{app:homotopy_variants}) to assign the remaining modes the smallest rescaled rate, ensuring contraction in every direction while retaining the Hessian eigenvectors. Energy Hessians and eigendecompositions are precomputed for the training data and reused for supervision; they are not supplied as network inputs, and generation requires no energy or Hessian evaluations. SemlaFlow integration and the homotopy variants are detailed in \cref{app:semla_integration,app:homotopy_variants}. Pseudocode for HI-FM training is provided in \cref{alg:hifm_loss}.

\subsection{Formal Analysis of HI-FM Conditional Paths}
\label{sec:hifm_properties}
Here, we confirm that while HI-FM's induced curvature changes the intermediate supervision, it does not affect neither the prior distribution nor the limiting target distribution. Therefore, importantly, when velocity targets and model predictions are square-integrable fields, the standard conditional-to-marginal FM regression identity still applies~\citep{Lipman2022FlowMF}. Let 
\[
    \mathbf{S}_t
    =\mathbf{P}\operatorname{diag}\!\left((1-t)^{r_i}\right)\mathbf{P}^{\top},
    \qquad
    \mathbf{K}_1
    =\mathbf{P}\operatorname{diag}(r_i)\mathbf{P}^{\top}.
\]

\newcommand{\hifmvaliditystatement}{%
Let $\mathbf{x}_0\sim\mathcal{N}(\mathbf{0},\mathbf{I})$ be independent
of $\mathbf{x}_1\sim q$. For each fixed $\mathbf{x}_1$ and
$t\in[0,1)$, the map in \cref{eq:hifm_sample_path_main} is invertible
and induces the conditional Gaussian in \cref{eq:hifm_conditional_gaussian}
with
\[
    \boldsymbol{\mu}_{\mathrm H}(t)=(\mathbf{I}-\mathbf{S}_t)\mathbf{x}_1,
    \qquad
    \boldsymbol{\Sigma}_{\mathrm H}(t)=\mathbf{S}_t^2.
\]
Its transporting conditional field is
\[
    \mathbf{u}^{\mathrm H}_t(\mathbf{x}\mid\mathbf{x}_1)
    =-\frac{\mathbf{K}_1(\mathbf{x}-\mathbf{x}_1)}{1-t}.
\]
}
\begin{proposition}[Valid transport and endpoint preservation]
\label[proposition]{prop:hifm_validity}
\hifmvaliditystatement
\end{proposition}

\begin{remark}
The path preserves $p_0$ at $t=0$ and converges to $\mathbf{x}_1$
as $t\rightarrow1$. Hence its marginal distribution converges to $q$.
If $r_i=1$ for all $i$, it recovers the linear conditional path.
\end{remark}
\emph{Proof sketch.} Apply the affine Gaussian transformation,
differentiate the sampling, and take its endpoint limit. The full proof is given in \cref{app:hifm_validity_proof}.

\par \section{Experiments}\paragraph{Evaluation logic.} Our experiments are structured to test both outcomes and mechanism. Matched data fractions and optimizer-update budgets ask whether the path-level change improves final molecular quality. Training trajectories then distinguish better final quality from faster optimization. Finally, Hessian--Jacobian alignment, path ablations, and an additional backbone test whether the effect reflects the proposed curvature-informed supervision rather than a particular schedule or architecture. 
\label{sec:experiments}

We evaluate whether this path-level change improves 3D molecular generation, a setting where the energy function and its Hessian have direct physical meaning.
Molecular conformations provide a natural setting for HI-FM: each
sample lies on a molecular energy landscape, and the Hessian at that
conformation describes which local motions are stiff or soft, i.e.,
which deformations increase the energy quickly or only weakly. We
consider Hessian-QM9 and GEOM-Drugs, and evaluate whether Hessian-informed
conditional paths improve generation quality, data efficiency, and
optimization efficiency.

\paragraph{Setup.}
After preprocessing,
Hessian-QM9 contains $30{,}974/6{,}356/4{,}151$
train/validation/test conformations.
For GEOM-Drugs, we retain molecules with at most $72$ atoms and split
the data by molecule into $80/10/10\%$ train/validation/test sets.
To evaluate data efficiency, we use $10\%$, $25\%$, $50\%$, and
$100\%$ subsets of the training data.

HI-FM is a conditional-path construction rather than a new
neural architecture. To isolate its contribution, we integrate HI-FM into SemlaFlow~\citep{irwin2024semlaflow} by
changing its conditional coordinate path and adapting its endpoint
regression target, as detailed in \cref{app:semla_integration}. We compare SemlaFlow trained with its linear coordinate path
against the same architecture trained with each of three
HI-FM path variants: direct (\hifmd{}), flow-time homotopy
(\hifmt{}), and epoch homotopy (\hifme{}), defined in
\cref{app:homotopy_variants}. Within each dataset and data fraction,
all methods use the same training subset, architecture, optimizer
settings, batching policy, and sampling procedure. We also match the
number of optimizer updates across methods and data fractions within
each dataset, so larger training sets do not receive additional
optimization steps. This design evaluates generation quality and data efficiency
under a matched optimization budget. Dataset preparation and training settings are detailed in
\cref{app:dataset_details,app:model_settings}. For Hessian-QM9, we use the supplied vacuum energy Hessians;
for GEOM-Drugs, we compute Cartesian Hessians of the MMFF94
energy (\cref{app:dataset_details}).
The main HI-FM experiments use zero diffusion, $c=2$, and
$\kappa=1$ (\cref{app:homotopy_variants}).

\paragraph{Evaluation.}
We use the same five main metrics on both datasets. Physical quality
is measured by MMFF94 energy and strain per atom, normalized by all
explicit atoms. In the endpoint tables, energy is averaged over
conformations with finite MMFF94 scores, while strain measures the
energy decrease after converged local relaxation. Chemical quality is
measured by molecular validity using generated bonds and formal
charges. Local geometric quality is measured by bond-length and
bond-angle Wasserstein-1 distances ($W_1$). Lower values are better
for all metrics except validity. Final generation results use the test sets. Sampling, uncertainty estimates, and metric
definitions are provided in
\cref{app:sampling_statistics,app:metrics}. Additional atom stability,
molecule stability, and uniqueness results are reported in
\cref{tab:semla_graph_quality}.

\subsection{Hessian-QM9}
\label{sec:qm9_results}

\paragraph{Data efficiency and generation quality.}
On Hessian-QM9, each model is trained for $33{,}000$ optimizer updates
using $10\%$, $25\%$, $50\%$, or $100\%$ of the training data.
The dataset preparation and training subsets are described in
\cref{app:qm9_semla_protocol}.

\begin{table*}[b!]
    \centering
    \scriptsize
    \setlength{\tabcolsep}{4pt}
    \caption{
        \textbf{SemlaFlow and SemlaFlow-HI-FM on Hessian-QM9.}
        Test results after $33{,}000$ optimizer updates.
        Energy and converged strain are in kcal\,mol$^{-1}$\,atom$^{-1}$.
        Values are mean$\pm$sample standard deviation over three
        training-seed means, each averaged over three generation replicates
        of $10{,}000$ molecules. Bold indicates the best mean within each
        data fraction.
    }
    \label{tab:qm9_semla}
    \resizebox{\textwidth}{!}{%
    \begin{tabular}{clccccc}
        \toprule
        & & \multicolumn{2}{c}{Physical quality}
        & \multicolumn{1}{c}{Chemical quality}
        & \multicolumn{2}{c}{Local geometry} \\
        \cmidrule(lr){3-4}
        \cmidrule(lr){5-5}
        \cmidrule(lr){6-7}
        Data & Method
        & Energy/atom $\downarrow$
        & Strain/atom $\downarrow$
        & Validity (\%) $\uparrow$
        & \shortstack{Bond-length $W_1$\\($10^{-3}$ \AA) $\downarrow$}
        & \shortstack{Bond-angle $W_1$\\($^\circ$) $\downarrow$} \\
        \midrule

        $10\%$ & SemlaFlow
        & $2.373 \pm 0.046$
        & $1.291 \pm 0.033$
        & $94.26 \pm 0.67$
        & $1.684 \pm 0.131$
        & $1.171 \pm 0.126$ \\

        $10\%$ & SemlaFlow-\hifmd{}
        & $\mathbf{2.098 \pm 0.017}$
        & $\mathbf{0.999 \pm 0.032}$
        & $\mathbf{94.86 \pm 0.50}$
        & $\mathbf{1.251 \pm 0.069}$
        & $\mathbf{0.962 \pm 0.160}$ \\

        $10\%$ & SemlaFlow-\hifmt{}
        & $2.143 \pm 0.069$
        & $1.020 \pm 0.047$
        & $94.43 \pm 0.08$
        & $1.266 \pm 0.089$
        & $1.080 \pm 0.193$ \\

        $10\%$ & SemlaFlow-\hifme{}
        & $2.167 \pm 0.029$
        & $1.058 \pm 0.035$
        & $94.81 \pm 0.15$
        & $1.319 \pm 0.074$
        & $0.995 \pm 0.170$ \\
        \midrule

        $25\%$ & SemlaFlow
        & $2.454 \pm 0.022$
        & $1.349 \pm 0.026$
        & $\mathbf{95.06 \pm 0.16}$
        & $1.569 \pm 0.131$
        & $1.190 \pm 0.147$ \\

        $25\%$ & SemlaFlow-\hifmd{}
        & $2.133 \pm 0.039$
        & $\mathbf{1.011 \pm 0.002}$
        & $94.77 \pm 0.45$
        & $1.242 \pm 0.077$
        & $1.050 \pm 0.149$ \\

        $25\%$ & SemlaFlow-\hifmt{}
        & $\mathbf{2.128 \pm 0.089}$
        & $1.034 \pm 0.088$
        & $94.73 \pm 0.20$
        & $\mathbf{1.241 \pm 0.108}$
        & $\mathbf{1.015 \pm 0.060}$ \\

        $25\%$ & SemlaFlow-\hifme{}
        & $2.153 \pm 0.020$
        & $1.021 \pm 0.028$
        & $94.90 \pm 0.20$
        & $1.281 \pm 0.037$
        & $1.109 \pm 0.136$ \\
        \midrule

        $50\%$ & SemlaFlow
        & $2.397 \pm 0.029$
        & $1.290 \pm 0.022$
        & $96.72 \pm 0.33$
        & $1.518 \pm 0.037$
        & $0.897 \pm 0.019$ \\

        $50\%$ & SemlaFlow-\hifmd{}
        & $\mathbf{2.232 \pm 0.118}$
        & $\mathbf{1.093 \pm 0.132}$
        & $96.91 \pm 0.28$
        & $1.137 \pm 0.128$
        & $\mathbf{0.729 \pm 0.032}$ \\

        $50\%$ & SemlaFlow-\hifmt{}
        & $2.264 \pm 0.099$
        & $1.144 \pm 0.085$
        & $96.81 \pm 0.26$
        & $\mathbf{1.129 \pm 0.092}$
        & $0.742 \pm 0.004$ \\

        $50\%$ & SemlaFlow-\hifme{}
        & $2.293 \pm 0.101$
        & $1.161 \pm 0.119$
        & $\mathbf{97.01 \pm 0.38}$
        & $1.273 \pm 0.089$
        & $0.797 \pm 0.051$ \\
        \midrule

        $100\%$ & SemlaFlow
        & $2.203 \pm 0.030$
        & $1.083 \pm 0.034$
        & $96.96 \pm 0.34$
        & $1.411 \pm 0.061$
        & $0.878 \pm 0.040$ \\

        $100\%$ & SemlaFlow-\hifmd{}
        & $\mathbf{2.046 \pm 0.007}$
        & $\mathbf{0.896 \pm 0.011}$
        & $\mathbf{97.04} \pm 0.04$
        & $\mathbf{1.083 \pm 0.068}$
        & $\mathbf{0.770 \pm 0.074}$ \\

        $100\%$ & SemlaFlow-\hifmt{}
        & $2.076 \pm 0.087$
        & $0.927 \pm 0.050$
        & $96.80 \pm 0.22$
        & $1.146 \pm 0.062$
        & $0.837 \pm 0.112$ \\

        $100\%$ & SemlaFlow-\hifme{}
        & $2.071 \pm 0.014$
        & $0.920 \pm 0.020$
        & $96.90 \pm 0.20$
        & $1.178 \pm 0.053$
        & $0.792 \pm 0.019$ \\
        \bottomrule
    \end{tabular}%
    }
\end{table*}

At every training-data fraction, all three HI-FM variants achieve
lower mean energy, strain, bond-length $W_1$, and bond-angle
$W_1$ than SemlaFlow trained on the same subset, with all models
receiving $33{,}000$ optimizer updates (\cref{tab:qm9_semla}).
With $100\%$ of the training data for both methods, the reductions
are $5.8$-$7.1\%$ for energy, $14.4$-$17.3\%$ for strain,
$16.5$-$23.2\%$ for bond-length $W_1$, and
$4.7$-$12.3\%$ for bond-angle $W_1$.
Validity remains comparable to SemlaFlow, with small changes
across methods and data fractions.

HI-FM also improves data efficiency substantially: all three variants trained
with only $10\%$ of the data achieve lower mean energy, strain,
and bond-length $W_1$ than SemlaFlow trained with $100\%$ of
the data, under the same optimizer-update budget.
In this $10\%$-versus-$100\%$ comparison, the reductions are
$1.6$-$4.8\%$ for energy, $2.3$-$7.8\%$ for strain, and
$6.5$-$11.3\%$ for bond-length $W_1$.
\Cref{fig:semla_energy_strain_compact}(a,b) shows this
cross-fraction advantage for energy and strain.

\paragraph{Interpretation.}
The comparison with full-data SemlaFlow suggests that local
energy curvature can compensate for fewer training conformations
on the evaluated energy, strain, and bond-length metrics,
under the same optimizer-update budget.
The improvements with the full dataset further show that
the benefit is not limited to settings where training data
are scarce.

\paragraph{Optimization efficiency.}
\Cref{fig:semla_energy_strain_compact}(e,f) shows the online
validation energy and strain per atom during training with $10\%$
of Hessian-QM9. All three
SemlaFlow-HI-FM variants reach or improve upon the best SemlaFlow value
observed over the full training run substantially earlier. For energy,
these crossings occur at epochs $2{,}170$-$2{,}300$, compared with
epoch $3{,}000$ for SemlaFlow; for strain, they occur at epochs
$2{,}040$-$2{,}120$, compared with epoch $2{,}970$. This corresponds
to approximately $23$-$28\%$ fewer optimizer updates for energy and
$29$-$31\%$ fewer for strain.

\subsection{GEOM-Drugs}
\label{sec:geom_drugs}

\paragraph{Data efficiency and generation quality.}
We next evaluate HI-FM on the larger and more diverse molecules in
GEOM-Drugs. Each model is trained for $1{,}045{,}600$ optimizer updates
using nested $10\%$, $25\%$, $50\%$, or $100\%$ molecule-level subsets
of the prepared training cohort. We follow the matched-optimization
design described above and report Exponential Moving Average (EMA) checkpoints throughout. Dataset preparation and the general setup are described in
\cref{app:geom_semla_protocol}.

\begin{table*}[t!]
    \centering
    \scriptsize
    \setlength{\tabcolsep}{4pt}
    \caption{
        \textbf{SemlaFlow and SemlaFlow-HI-FM on GEOM-Drugs.}
        Test results using final EMA weights after $1{,}045{,}600$ optimizer updates.
        Energy and converged strain are in kcal\,mol$^{-1}$\,atom$^{-1}$.
        Values are mean$\pm$population standard deviation over three
        generation replicates of $10{,}000$ molecules from one training
        seed. Bold indicates the best mean within each data fraction.
    }
    \label{tab:geom_semla}
    \resizebox{\textwidth}{!}{%
    \begin{tabular}{clccccc}
        \toprule
        & & \multicolumn{2}{c}{Physical quality}
        & \multicolumn{1}{c}{Chemical quality}
        & \multicolumn{2}{c}{Local geometry} \\
        \cmidrule(lr){3-4}
        \cmidrule(lr){5-5}
        \cmidrule(lr){6-7}
        Data & Method
        & Energy/atom $\downarrow$
        & Strain/atom $\downarrow$
        & Validity (\%) $\uparrow$
        & \shortstack{Bond-length $W_1$\\($10^{-3}$ \AA) $\downarrow$}
        & \shortstack{Bond-angle $W_1$\\($^\circ$) $\downarrow$} \\
        \midrule

        $10\%$ & SemlaFlow
        & $2.060 \pm 0.020$
        & $1.318 \pm 0.016$
        & $90.47 \pm 0.13$
        & $0.813 \pm 0.005$
        & $1.640 \pm 0.012$ \\

        $10\%$ & SemlaFlow-\hifmd{}
        & $1.783 \pm 0.006$
        & $1.081 \pm 0.014$
        & $\mathbf{91.49 \pm 0.19}$
        & $0.603 \pm 0.005$
        & $\mathbf{1.638 \pm 0.011}$ \\

        $10\%$ & SemlaFlow-\hifmt{}
        & $\mathbf{1.762 \pm 0.023}$
        & $\mathbf{1.044 \pm 0.008}$
        & $90.70 \pm 0.09$
        & $0.660 \pm 0.000$
        & $1.650 \pm 0.011$ \\

        $10\%$ & SemlaFlow-\hifme{}
        & $1.855 \pm 0.005$
        & $1.082 \pm 0.001$
        & $90.66 \pm 0.29$
        & $\mathbf{0.593 \pm 0.005}$
        & $1.648 \pm 0.015$ \\
        \midrule

        $25\%$ & SemlaFlow
        & $2.069 \pm 0.014$
        & $1.336 \pm 0.015$
        & $92.70 \pm 0.33$
        & $0.853 \pm 0.005$
        & $1.684 \pm 0.004$ \\

        $25\%$ & SemlaFlow-\hifmd{}
        & $1.880 \pm 0.018$
        & $1.176 \pm 0.019$
        & $\mathbf{92.99 \pm 0.24}$
        & $0.653 \pm 0.005$
        & $1.786 \pm 0.019$ \\

        $25\%$ & SemlaFlow-\hifmt{}
        & $\mathbf{1.801 \pm 0.005}$
        & $\mathbf{1.066 \pm 0.003}$
        & $92.97 \pm 0.09$
        & $0.660 \pm 0.000$
        & $\mathbf{1.570 \pm 0.018}$ \\

        $25\%$ & SemlaFlow-\hifme{}
        & $1.863 \pm 0.010$
        & $1.119 \pm 0.003$
        & $92.53 \pm 0.33$
        & $\mathbf{0.623 \pm 0.005}$
        & $1.623 \pm 0.004$ \\
        \midrule

        $50\%$ & SemlaFlow
        & $2.097 \pm 0.016$
        & $1.356 \pm 0.016$
        & $93.29 \pm 0.14$
        & $0.770 \pm 0.000$
        & $\mathbf{1.660 \pm 0.010}$ \\

        $50\%$ & SemlaFlow-\hifmd{}
        & $1.943 \pm 0.014$
        & $1.216 \pm 0.006$
        & $94.44 \pm 0.28$
        & $0.647 \pm 0.005$
        & $1.790 \pm 0.008$ \\

        $50\%$ & SemlaFlow-\hifmt{}
        & $1.906 \pm 0.005$
        & $1.191 \pm 0.000$
        & $\mathbf{94.48 \pm 0.20}$
        & $\mathbf{0.603 \pm 0.005}$
        & $1.689 \pm 0.002$ \\

        $50\%$ & SemlaFlow-\hifme{}
        & $\mathbf{1.854 \pm 0.011}$
        & $\mathbf{1.155 \pm 0.004}$
        & $94.20 \pm 0.09$
        & $0.673 \pm 0.005$
        & $1.761 \pm 0.017$ \\
        \midrule

        $100\%$ & SemlaFlow
        & $2.012 \pm 0.018$
        & $1.263 \pm 0.006$
        & $94.33 \pm 0.24$
        & $0.783 \pm 0.009$
        & $1.649 \pm 0.005$ \\

        $100\%$ & SemlaFlow-\hifmd{}
        & $\mathbf{1.808 \pm 0.017}$
        & $\mathbf{1.074 \pm 0.002}$
        & $93.97 \pm 0.12$
        & $\mathbf{0.557 \pm 0.012}$
        & $1.470 \pm 0.017$ \\

        $100\%$ & SemlaFlow-\hifmt{}
        & $1.865 \pm 0.024$
        & $1.142 \pm 0.016$
        & $93.49 \pm 0.33$
        & $0.583 \pm 0.005$
        & $\mathbf{1.457 \pm 0.013}$ \\

        $100\%$ & SemlaFlow-\hifme{}
        & $1.812 \pm 0.006$
        & $1.098 \pm 0.006$
        & $\mathbf{94.57 \pm 0.20}$
        & $0.627 \pm 0.005$
        & $1.594 \pm 0.010$ \\
        \bottomrule
    \end{tabular}%
    }
\end{table*}

\begin{figure*}[t]
    \centering
    \includegraphics[width=\textwidth]
        {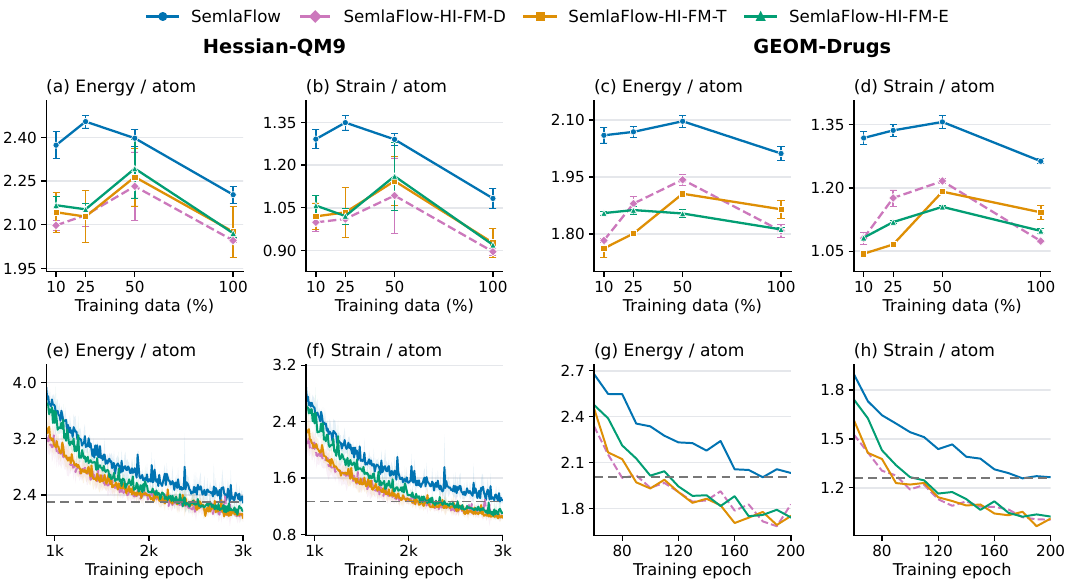}
    \caption{
    \textbf{Physical quality and optimization efficiency on Hessian-QM9
    and GEOM-Drugs.}
    \textbf{(a--d)} Final energy and strain per atom across training-data
    fractions.
    \textbf{(e--h)} Energy and strain during training with $10\%$ data;
    dashed lines mark the corresponding SemlaFlow reference values.
    Hessian-QM9 shows mean$\pm$sample standard deviation over three
    training seeds. GEOM-Drugs endpoint error bars show population
    standard deviation over three generation replicates from one seed;
    training curves show the single trajectory. Lower is better.
    }
    \label{fig:semla_energy_strain_compact}
\end{figure*}

\Cref{tab:geom_semla} shows consistent improvements when models are
trained with the same amount of data. At every training-data fraction,
all three HI-FM variants reduce mean energy, strain, and bond-length
$W_1$ relative to SemlaFlow trained on the same subset. Across the
twelve comparisons, the reductions are $7.3$-$14.5\%$ for energy,
$9.6$-$20.8\%$ for strain, and $12.6$-$28.9\%$ for bond-length
$W_1$. \Cref{fig:semla_energy_strain_compact}(c,d) shows the corresponding
separation in energy and strain across all four data fractions. Overall, \hifmd{}, the simplest, is best on most Hessian-QM9 metrics while the homotopies are best at several smaller GEOM-Drugs fractions.

With the full training set, HI-FM reduces energy by $7.3$-$10.1\%$,
strain by $9.6$-$15.0\%$, and bond-length $W_1$ by
$19.9$-$28.9\%$ relative to full-data SemlaFlow. All three variants
also improve bond-angle $W_1$ in this setting, with reductions of
$3.3$-$11.6\%$. Bond-angle changes are mixed at smaller data
fractions, while validity remains broadly comparable and improves
in most matched-data comparisons.

\paragraph{Interpretation.}
The GEOM-Drugs results show that the improvements in energy,
strain, and bond-length geometry extend beyond the smaller
molecules in Hessian-QM9.
These gains are obtained using MMFF94 Hessians, supporting
the use of force-field curvature to shape the conditional
paths for larger and more diverse molecules.

\paragraph{Optimization efficiency.}
\Cref{fig:semla_energy_strain_compact}(g,h) compares the energy and
strain trajectories when all methods use the same $10\%$ training subset.
All three HI-FM variants reach the matched SemlaFlow baseline's
final energy and strain earlier in the recorded training history.
Their advantage is visible before the final checkpoint and persists
across much of the later trajectory. 

\subsection{Additional Analyses}
\label{sec:additional_analyses}

\paragraph{Hessian-Jacobian alignment.}
We test whether HI-FM's learned vector fields reflect the local
energy curvature used to construct the conditional paths.
We compare their Jacobians with the energy Hessians near the
data endpoint using ProjSim@10 for directional alignment and
HessianFit for matrix alignment, defined in
\cref{app:hjac_alignment}.
\Cref{tab:hjac_main} reports results for models trained on the
full training sets, using identical evaluation coordinates
across methods.
All three HI-FM variants achieve higher mean alignment than
SemlaFlow on both metrics and datasets.
The same pattern holds across all evaluated data fractions
and both probe conditions (\cref{app:hjac_alignment}). A qualitative C=O-stretching example is shown in \cref{fig:co_stretch_alignment} in the appendix.

\begin{table}[t]
    \centering
    \footnotesize
    \setlength{\tabcolsep}{4pt}
    \caption{
        \textbf{Hessian-Jacobian alignment with full training data.}
        Final EMA models are evaluated at shared-linear probes near
        $t=1$, using $500$ validation molecules per dataset.
        Hessian-QM9 reports mean$\pm$sample SD across three
        training-seed means; GEOM-Drugs reports mean$\pm$population
        SD across molecules for one training seed.
        Higher is better; bold marks the best mean in each column.
    }
    \label{tab:hjac_main}
    \begin{tabular}{@{}lcccc@{}}
        \toprule
        & \multicolumn{2}{c}{Hessian-QM9}
        & \multicolumn{2}{c}{GEOM-Drugs} \\
        \cmidrule(lr){2-3}\cmidrule(lr){4-5}
        Method
        & ProjSim@10 $\uparrow$ & HessianFit $\uparrow$
        & ProjSim@10 $\uparrow$ & HessianFit $\uparrow$ \\
        \midrule

        SemlaFlow
        & $0.3155 \pm 0.0118$ & $0.7046 \pm 0.0014$
        & $0.1148 \pm 0.0590$ & $0.6893 \pm 0.0110$ \\

        SemlaFlow-\hifmd{}
        & $0.7871 \pm 0.0104$ & $0.7716 \pm 0.0021$
        & $0.6033 \pm 0.0995$ & $0.7621 \pm 0.0117$ \\

        SemlaFlow-\hifmt{}
        & $\mathbf{0.7972 \pm 0.0042}$
        & $\mathbf{0.7846 \pm 0.0012}$
        & $\mathbf{0.6473 \pm 0.0882}$
        & $\mathbf{0.7767 \pm 0.0104}$ \\

        SemlaFlow-\hifme{}
        & $0.7436 \pm 0.0207$ & $0.7622 \pm 0.0039$
        & $0.6294 \pm 0.0964$ & $0.7671 \pm 0.0121$ \\

        \bottomrule
    \end{tabular}
\end{table}

\paragraph{Ablation studies.}
On $10\%$ of Hessian-QM9, randomizing active Hessian eigendirections
while preserving the spectrum removes approximately $76\%$ of
direct HI-FM's energy gain and $72\%$ of its strain gain over
SemlaFlow (\cref{app:directional_control}), indicating that the
target-specific eigendirections contribute beyond the eigenvalue
spectrum alone.
Sensitivity to $c$ and $\kappa$ is reported in
\cref{app:hyperparameter_sensitivity}.

\paragraph{Additional backbone.}
With an EDM-style EGNN~\citep{Satorras2021EGNN,Hoogeboom2022EDM} trained by flow matching, all three HI-FM
variants reduce seed-averaged median energy and strain per atom
and improve mean validity across all evaluated Hessian-QM9 data
fractions relative to the same-backbone OT-FM baseline.
Across these twelve matched-data comparisons, energy is reduced
by $4.9$-$12.7\%$ and strain by $10.1$-$18.2\%$
(\cref{app:qm9_egnn}).

\paragraph{Computational overhead.}
With cached eigensystems, direct HI-FM increases time per optimizer
update by $18.9\%$ relative to SemlaFlow (\cref{app:computational_overhead}).
Generation requires no energy or Hessian evaluations.

\section{Conclusion}
In this paper, we investigated the utility of using conditional
flows in the flow matching framework to incorporate local
characteristics of the energy landscape of a target distribution.
To do this, we used a linear time-invariant stochastic differential
equation based on the Hessian of an energy function, and transformed
its conditional probability paths to obtain a consistent sampling
and integration interval. In the process, we connected flow
matching with local linearization and group invariance.
Our experiments with SemlaFlow on Hessian-QM9 and GEOM-Drugs show
that these Hessian-informed paths improve the physical and
geometric quality of generated molecules and their data efficiency
under matched optimizer-update budgets.
These results show that incorporating local energy landscape
information into flow matching provides a useful inductive bias,
without requiring energy or Hessian evaluations during generation.
\paragraph{Limitations.}
HI-FM introduces an additional preprocessing cost through the
computation and storage of target energy Hessians and their
eigen decompositions. These quantities are computed once offline
and reused during training, while generation requires no energy or Hessian evaluations. For GEOM-Drugs, we use MMFF94 rather than DFT Hessians, so the
conditional paths reflect the local curvature of the chosen
force field.

\section*{AI Use Statement}

In this work, we used generative AI tools for manuscript editing and
language polishing, including improvements to grammar, wording, and
readability and rephrasing the formal statement and its proof. We have reviewed all AI-assisted work. We take responsibility
for the final content of this work, including text, claims, and artifacts
produced with the aid of generative AI.

\section*{Reproducibility Statement}

We describe HI-FM in \cref{sec:hifm} and provide pseudocode
in \cref{alg:hifm_loss}, with the full derivation and its
assumptions given in \cref{app:hifm_derivation}.
Dataset preparation, model architectures, training
hyperparameters, and sampling and evaluation procedures
are detailed in \cref{app:implementation_details}.
Protocols for the additional backbone, Hessian-Jacobian
analysis, and ablation studies are provided in
\cref{app:qm9_egnn,app:hjac_alignment,app:ablations}.

\bibliography{refs}
\bibliographystyle{iclr2027_conference}

\newpage
\appendix

\section{HI-FM Derivation}\label{app:hifm_derivation}\subsection{Derivation of Hessian-Informed Conditional Paths}\label{sec:interpolant}\label{sec:topological}We derive how the energy Hessian determines the conditional mean, covariance, and velocity field used by HI-FM. Our construction defines conditional dynamics through the target data Hessian and maps their relaxation to finite flow time. The Gaussian distributions and transporting fields follow by applying standard results for linear stochastic systems and probability-flow ODEs to this construction. We first describe energy relaxation, from which we construct the conditional paths. To include both motion driven by the energy gradient and additive fluctuations, consider\begin{equation}    \mathrm{d}\mathbf{X}_s=-\nabla_{\mathbf{x}}V(\mathbf{X}_s)\,\mathrm{d}s+\mathbf{B}_1\,\mathrm{d}\mathbf{W}_s,    \label{eq:hifm_energy_sde}\end{equation}where $s\geq0$ denotes relaxation time, $\mathbf{X}_s\in\mathbb{R}^d$ is the random state, and $\mathbf{W}_s$ is standard $d$-dimensional Brownian motion. The constant diffusion matrix $\mathbf{B}_1\in\mathbb{R}^{d\times d}$ controls the magnitude and directional structure of the random fluctuations. We use the local linearization of this energy model to construct conditional distributions, not as the learned generative process. A deterministic field transporting these distributions will provide the FM velocity supervision. For the derivation, assume that $V$ is twice continuously differentiable and $\mathbf{x}_1$ is a non-degenerate minimum, so that $\nabla V(\mathbf{x}_1)=\mathbf{0}$ and all $\lambda_i>0$; we address molecular null modes below. Linearizing the drift gives\begin{equation}    -\nabla_{\mathbf{x}}V(\mathbf{x})\approx-\mathbf{H}_1(\mathbf{x}-\mathbf{x}_1).    \label{eq:hifm_local_linearization}\end{equation}For zero diffusion, the Hartman-Grobman theorem establishes local topological conjugacy between the deterministic gradient flow and this linearization~\citep{hartman1960lemma}, giving a local correspondence between their trajectories. For HI-FM, we use this target-specific approximation to define tractable conditional dynamics:\begin{equation}    \mathrm{d}\mathbf{X}_s=-\mathbf{H}_1(\mathbf{X}_s-\mathbf{x}_1)\,\mathrm{d}s+\mathbf{B}_1\,\mathrm{d}\mathbf{W}_s,\qquad \mathbf{B}_1=\mathbf{P}\operatorname{diag}(\beta_i)\mathbf{P}^{\top},    \label{eq:hifm_linear_sde}\end{equation}where $\beta_i\geq0$ is the diffusion amplitude along Hessian eigenvector $i$. Choosing $\mathbf{B}_1$ in the same eigenbasis makes the dynamics decouple along these directions. The following Gaussian formulas are exact for this linear model; its correspondence with the nonlinear energy dynamics is local. Let the Gaussian source $\mathbf{X}_0\sim p_0$ have mean $\boldsymbol{\mu}_0$ and covariance $\boldsymbol{\Sigma}_0=\mathbf{P}\operatorname{diag}(\sigma_{0,i}^2)\mathbf{P}^{\top}$, where $\sigma_{0,i}^2>0$ is the source variance along eigenvector $i$, and assume independence from the Brownian motion. The standard Gaussian source satisfies this covariance assumption for every target, with $\boldsymbol{\mu}_0=\mathbf{0}$ and $\sigma_{0,i}^2=1$. Applying the standard linear-SDE solution \citep[Section~6.2]{Srkk2019AppliedSD} gives the conditional mean $\boldsymbol{\mu}(s)$ and covariance $\boldsymbol{\Sigma}(s)$:\begin{equation}\begin{aligned}    \boldsymbol{\mu}(s)&=\mathbf{x}_1+\mathbf{P}\operatorname{diag}\!\left(e^{-\lambda_i s}\right)\mathbf{P}^{\top}(\boldsymbol{\mu}_0-\mathbf{x}_1),\\    \boldsymbol{\Sigma}(s)&=\mathbf{P}\operatorname{diag}\!\left[\frac{\beta_i^2}{2\lambda_i}+e^{-2\lambda_i s}\left(\sigma_{0,i}^2-\frac{\beta_i^2}{2\lambda_i}\right)\right]\mathbf{P}^{\top}.\end{aligned}\label{eq:probability_path_time}\end{equation}The Hessian eigenvalues determine the mean's relaxation rates, while each conditional variance approaches $\beta_i^2/(2\lambda_i)$. Thus, diffusion permits nonzero terminal spread, and setting all $\beta_i=0$ gives a point target. To obtain deterministic velocity supervision for these Hessian-based distributions, we apply the probability-flow construction~\citep{Song2020ScoreBasedGM,Maoutsa2020InteractingPS}. Substituting the Gaussian score $-\boldsymbol{\Sigma}(s)^{-1}(\mathbf{x}-\boldsymbol{\mu}(s))$ and the linear drift gives\begin{equation}    \mathbf{w}_s(\mathbf{x}\mid\mathbf{x}_1)=-\mathbf{H}_1(\mathbf{x}-\mathbf{x}_1)+\frac{1}{2}\mathbf{B}_1\mathbf{B}_1^{\top}\boldsymbol{\Sigma}(s)^{-1}(\mathbf{x}-\boldsymbol{\mu}(s)),    \label{eq:linear_pfvf}\end{equation}where $\mathbf{w}_s$ is the conditional field in relaxation time. Its ODE transports the same conditional distributions as \cref{eq:hifm_linear_sde}, not its stochastic trajectories. The score is available analytically; no score network is trained. The relaxation approaches its limiting distribution as $s\to\infty$, whereas our FM construction uses a common finite interval. Let $\lambda_{\min}=\min_i\lambda_i>0$ be the slowest relaxation rate. It bounds the displacement of the conditional mean:\begin{equation}    \|\boldsymbol{\mu}(s)-\mathbf{x}_1\|_2\leq e^{-\lambda_{\min}s}\|\boldsymbol{\mu}_0-\mathbf{x}_1\|_2.    \label{eq:metric}\end{equation}For HI-FM, we define flow time using this exponential envelope, rather than the exact mean distance, to obtain an explicitly invertible clock:\begin{equation}    t=1-e^{-\kappa\lambda_{\min}s},\qquad s(t)=-\frac{\log(1-t)}{\kappa\lambda_{\min}},\qquad r_i=\frac{\lambda_i}{\kappa\lambda_{\min}},    \label{eq:hifm_clock}\end{equation}where $\kappa>0$ controls the time scaling and $r_i$ is the resulting exponent along eigenvector $i$. This maps $s\in[0,\infty)$ to $t\in[0,1)$ and gives $e^{-\lambda_i s(t)}=(1-t)^{r_i}$. The exponents encode relative relaxation rates. For $\kappa=1$, the mean displacement along the slowest Hessian direction decreases by the factor $1-t$. Substituting $s(t)$ from \cref{eq:hifm_clock} into \cref{eq:probability_path_time} gives the finite-time HI-FM conditional moments:\begin{equation}\begin{aligned}    \boldsymbol{\mu}_{\mathrm H}(t)&=\mathbf{x}_1+\mathbf{P}\operatorname{diag}\!\left((1-t)^{r_i}\right)\mathbf{P}^{\top}(\boldsymbol{\mu}_0-\mathbf{x}_1),\\    \boldsymbol{\Sigma}_{\mathrm H}(t)&=\mathbf{P}\operatorname{diag}\!\left[\frac{\beta_i^2}{2\lambda_i}+(1-t)^{2r_i}\left(\sigma_{0,i}^2-\frac{\beta_i^2}{2\lambda_i}\right)\right]\mathbf{P}^{\top}.\end{aligned}\label{eq:probability_path}\end{equation}To obtain its velocity target for FM, we apply the same time transformation to the conditional field $\mathbf{w}_s$ in \cref{eq:linear_pfvf}. The chain rule gives\begin{equation}    \mathbf{u}^{\mathrm H}_t(\mathbf{x}\mid\mathbf{x}_1)=\mathbf{w}_{s(t)}(\mathbf{x}\mid\mathbf{x}_1)\frac{\mathrm{d}s}{\mathrm{d}t}=\frac{\mathbf{w}_{s(t)}(\mathbf{x}\mid\mathbf{x}_1)}{\kappa\lambda_{\min}(1-t)},\qquad t\in[0,1).    \label{eq:hifm_finite_field}\end{equation}Thus, $\mathbf{u}^{\mathrm H}_t$ transports $p_t^{\mathrm H}$ and provides the conditional velocity supervision in \cref{eq:fm}.For zero diffusion, the conditional sampling map is\begin{equation}    \mathbf{x}^{\mathrm H}_t=\mathbf{x}_1+\mathbf{P}\operatorname{diag}\!\left((1-t)^{r_i}\right)\mathbf{P}^{\top}(\mathbf{x}_0-\mathbf{x}_1),\qquad \mathbf{x}_0\sim p_0.    \label{eq:hifm_sample_path}\end{equation}Its time derivative equals $\mathbf{u}^{\mathrm H}_t(\mathbf{x}^{\mathrm H}_t\mid\mathbf{x}_1)$. At $t=0$, the map is the identity and therefore preserves $p_0$. Since each $r_i>0$, it converges to $\mathbf{x}_1$ as $t\to1$. With $\kappa=1$, equal eigenvalues recover the linear conditional path in \cref{eq:linear_conditional_path}.The derivation above assumes a positive-definite Hessian. For molecular energies invariant to translations and rotations, however, infinitesimal rigid-body motions lie in the Hessian nullspace at equilibrium and do not contract under the linearized drift. We rescale eigenvalues above a positive threshold to control the spread of relaxation rates. For the remaining modes, we use \emph{hyperbolization} (\cref{app:homotopy_variants}) to assign the smallest rescaled rate, ensuring contraction in every direction. We use the processed rates $\alpha_i$ and their minimum in place of $\lambda_i$ and $\lambda_{\min}$ throughout the path construction, with $\mathbf{H}_1$ replaced by the corresponding matrix reconstructed in the same eigenbasis. This retains the Hessian eigenvectors but modifies the relaxation dynamics. The conditional samples and velocities above provide the supervision in \cref{eq:fm}, without simulating the relaxation. For SemlaFlow, we adapt the coordinate regression target while retaining its endpoint parameterization and time sampling; the corresponding time-weighted velocity regression is detailed in \cref{app:semla_integration}. Energy Hessians of target data and eigendecompositions are precomputed and reused for supervision, not supplied as additional network inputs; generation requires no energy or Hessian evaluations. Alongside direct HI-FM (\hifmd{}), we evaluate flow-time (\hifmt{}) and epoch (\hifme{}) homotopy variants, defined in \cref{app:homotopy_variants}.

\subsection{Proof of \cref{prop:hifm_validity}}
\label{app:hifm_properties_proofs}
\label{app:hifm_validity_proof}
We restate and prove the results from \cref{sec:hifm_properties}. Recall that
\[
\begin{gathered}
    \mathbf{H}_1=\mathbf{P}\operatorname{diag}(\lambda_i)\mathbf{P}^{\top},
    \qquad
    r_i=\frac{\alpha_i}{\kappa\alpha_{\min}}>0,
    \qquad
    \alpha_{\min}=\min_i\alpha_i,\\
    \mathbf{S}_t=\mathbf{P}\operatorname{diag}\!\left((1-t)^{r_i}\right)
                    \mathbf{P}^{\top},
    \qquad
    \mathbf{K}_1=\mathbf{P}\operatorname{diag}(r_i)\mathbf{P}^{\top}.
\end{gathered}
\] Capital letters $\mathbf{X}_0$, $\mathbf{X}_1$, and $\mathbf{X}^{\mathrm H}_t$
below denote the random variables corresponding to the sampled points.

\renewcommand{\hifmrestatedname}{Proposition~\ref{prop:hifm_validity}}
\begin{hifmrepeatedproposition}[Valid transport and endpoint preservation]
\hifmvaliditystatement
\end{hifmrepeatedproposition}

\begin{proof}
\textbf{Step 1: The sampling map is an invertible affine map.}
Fix $\mathbf{x}_1$ and $t<1$. Since $(1-t)^{r_i}>0$ for every $i$,
$\mathbf{S}_t$ is symmetric positive definite. Thus
\[
    \mathbf{S}_t^{-1}
    =\mathbf{P}\operatorname{diag}\!\left((1-t)^{-r_i}\right)\mathbf{P}^{\top},
    \qquad
    \det\mathbf{S}_t=(1-t)^{\sum_i r_i}>0.
\]
The inverse of $\mathbf{x}=\mathbf{x}_1+\mathbf{S}_t(\mathbf{x}_0-\mathbf{x}_1)$
is
\[
    \mathbf{x}_0=\mathbf{x}_1+\mathbf{S}_t^{-1}(\mathbf{x}-\mathbf{x}_1).
\]
The map is a diffeomorphism for $t<1$.

\medskip\noindent
\textbf{Step 2: Apply the affine transformation rule for Gaussians.}
Conditionally on $\mathbf{X}_1=\mathbf{x}_1$, independence leaves
$\mathbf{X}_0\sim\mathcal{N}(\mathbf{0},\mathbf{I})$, and
\[
    \mathbf{X}^{\mathrm H}_t
    =(\mathbf{I}-\mathbf{S}_t)\mathbf{x}_1+\mathbf{S}_t\mathbf{X}_0.
\]
Taking the conditional mean and covariance gives
\begin{align*}
    \mathbb{E}[\mathbf{X}^{\mathrm H}_t\mid\mathbf{X}_1=\mathbf{x}_1]
    &=(\mathbf{I}-\mathbf{S}_t)\mathbf{x}_1,\\
    \operatorname{Cov}(\mathbf{X}^{\mathrm H}_t\mid\mathbf{X}_1=\mathbf{x}_1)
    &=\mathbf{S}_t\mathbf{I}\mathbf{S}_t^{\top}=\mathbf{S}_t^2.
\end{align*}
These are the moments of the affine Gaussian output.

\medskip\noindent
\textbf{Step 3: Differentiate the map to obtain its transporting field.}
Differentiation with the target and source fixed yields
\begin{equation}
\begin{aligned}
    \partial_t\mathbf{S}_t
    &=-\mathbf{P}\operatorname{diag}\!\left(r_i(1-t)^{r_i-1}\right)
         \mathbf{P}^{\top}
      =-\frac{\mathbf{K}_1\mathbf{S}_t}{1-t},\\
    \partial_t\mathbf{x}^{\mathrm H}_t
    &=(\partial_t\mathbf{S}_t)(\mathbf{x}_0-\mathbf{x}_1)
      =-\frac{\mathbf{K}_1(\mathbf{x}^{\mathrm H}_t-\mathbf{x}_1)}{1-t}.
\end{aligned}
\label{eq:hifm_proof_velocity}
\end{equation}
We can then replace $\mathbf{x}^{\mathrm H}_t$ by any current
position $\mathbf{x}$. On each $[0,T]$ with $T<1$, the field is continuous
in time and globally Lipschitz in $\mathbf{x}$. Its unique ODE flow is
therefore the sampling map, which transports $p_0$ to the Gaussian in Step 2.

\medskip\noindent
\textbf{Step 4: Check both endpoints and the marginal limit.}
At $t=0$, $\mathbf{S}_0=\mathbf{I}$ and
$\mathbf{x}^{\mathrm H}_0=\mathbf{x}_0$.
For each fixed target, let $r_{\min}=\min_i r_i>0$. Orthogonality gives
\[
    \|\mathbf{x}^{\mathrm H}_t-\mathbf{x}_1\|_2
    \leq (1-t)^{r_{\min}}\|\mathbf{x}_0-\mathbf{x}_1\|_2
    \longrightarrow0.
\]
Under the joint sampling of $(\mathbf{X}_0,\mathbf{X}_1)$ this convergence
holds:
\[
    \lim_{t\rightarrow1}\mathbb{E}[f(\mathbf{X}^{\mathrm H}_t)]
    =\mathbb{E}[f(\mathbf{X}_1)].
\]
The same argument with a fixed target gives conditional convergence to
$\delta_{\mathbf{x}_1}$.
Finally, when every $r_i=1$, $\mathbf{S}_t=(1-t)\mathbf{I}$, so
\[
    \mathbf{x}^{\mathrm H}_t
    =\mathbf{x}_1+(1-t)(\mathbf{x}_0-\mathbf{x}_1)
    =(1-t)\mathbf{x}_0+t\mathbf{x}_1,
\]
which is the linear conditional path.
\end{proof}

\section{Implementation Details}
\label{app:implementation_details}

\subsection{Datasets and Preprocessing}
\label{app:dataset_details}

\subsubsection{Hessian-QM9}
\label{app:qm9_semla_protocol}

We use Hessian-QM9~\citep{williams2025hessian} and the vacuum
Hessian provided for each molecule. We retain explicit hydrogen atoms and
align the molecular graph and formal charges with the corresponding
coordinate and Hessian ordering. The training, validation, and test sets
contain $30{,}974$, $6{,}356$, and $4{,}151$ conformations,
respectively, with no shared chemical identities across splits.
For the data-efficiency experiments, we use nested $10\%$, $25\%$,
$50\%$, and $100\%$ training subsets containing $3{,}100$, $7{,}743$,
$15{,}491$, and $30{,}974$ conformations.

\subsubsection{GEOM-Drugs}
\label{app:geom_semla_protocol}

We use molecular conformations from GEOM-Drugs~\citep{axelrod2022geom},
retaining molecules with at most $72$ atoms including hydrogens. For
the retained conformations, we additionally compute Cartesian energy
Hessians at the corresponding molecular geometries. 
Given Cartesian coordinates $\mathbf{x} \in \mathbb{R}^{3N}$, we computed the Hessians using the MMFF94 force field~\citep{Halgren1996}, reimplemented in PyTorch to enable automatic differentiation. Denoting the MMFF94 potential energy as $E(\mathbf{x})$, the Hessian can be computed as follows.
\begin{equation}
\mathbf{H}(\mathbf{x}) \;=\; \nabla^{2} E(\mathbf{x}) \;\in\; \mathbb{R}^{3N \times 3N},
\qquad
H_{ij}(\mathbf{x}) \;=\; \frac{\partial^{2} E(\mathbf{x})}{\partial x_i \, \partial x_j},
\end{equation}

We split the data by molecular identity into $80/10/10\%$
training, validation, and test sets, and retain all conformations
associated with each selected identity. The $10\%$, $25\%$, $50\%$,
and $100\%$ training sets are nested subsets of the same training
split. The validation and test sets contain $563{,}401$ and
$565{,}497$ conformations, respectively, corresponding to $23{,}957$
molecular identities in each split, and remain fixed across all
experiments.

\subsection{Architecture and Training}
\label{app:model_settings}

We use SemlaFlow~\citep{irwin2024semlaflow} as the backbone for
the main experiments. The baseline and HI-FM variants use the same architecture,
optimizer, batching, and auxiliary objectives; HI-FM only changes the
coordinate path and its corresponding training target. We retain
SemlaFlow's self-conditioning and EMA weights, while disabling
coordinate optimal-transport coupling and additional coordinate-path
noise for all methods. This keeps the source-target pairing and
coordinate perturbation identical across methods, so that the comparison
primarily reflects the effect of replacing the linear coordinate path
with the Hessian-informed one. The SemlaFlow settings are summarized in
\cref{tab:semla_model_training}. We also evaluate an EDM-style
EGNN trained with flow matching on Hessian-QM9; its architecture
and training settings are given in \cref{app:qm9_egnn}. 

\paragraph{Computational resources.}
The main experiments used NVIDIA A100 GPUs.
Hessian-QM9 training-and-evaluation jobs averaged approximately
7 hours on one GPU per model (7 GPU-hours). GEOM-Drugs training across both stages took
approximately 50 hours on four GPUs per model (200 GPU-hours), including online validation.

\begin{table*}[b]
\centering
\footnotesize
\setlength{\tabcolsep}{6pt}
\caption{SemlaFlow architecture and training settings for Hessian-QM9
and GEOM-Drugs. Shared entries span both datasets.}
\label{tab:semla_model_training}
\begin{tabular*}{\textwidth}{@{\extracolsep{\fill}}lll@{}}
\toprule
Setting & Hessian-QM9 & GEOM-Drugs \\
\midrule

Hidden width / layers
& \multicolumn{2}{c}{$384/12$} \\

Message / hidden-message / edge width
& \multicolumn{2}{c}{$128/128/128$} \\

Coordinate sets / attention heads
& \multicolumn{2}{c}{$64/32$} \\

Global atom-cost batch budget
& \multicolumn{2}{c}{$4{,}096$} \\

Optimizer
& \multicolumn{2}{c}{Adam with AMSGrad} \\

Learning rate / weight decay
& \multicolumn{2}{c}{$3\times10^{-4}/0$} \\

Gradient clipping
& \multicolumn{2}{c}{$1.0$} \\

Sampled flow time
& \multicolumn{2}{c}{$\mathrm{Beta}(2,1)$} \\

Self-conditioning probability / EMA decay
& \multicolumn{2}{c}{$0.5/0.999$} \\

\midrule
GPUs per model
& $1$ & $4$ \\

Linear LR warm-up
& $2{,}000$ updates & $10{,}000$ updates \\

Coordinate / atom / bond / charge loss weights
& $1/0.2/0.5/1$ & $1/0.2/1/1$ \\

Optimizer updates
& $33{,}000$ & $1{,}045{,}600$ \\

\bottomrule
\end{tabular*}
\end{table*}

\begin{algorithm}[t!]
\caption{Hessian-Informed Flow Matching}
\label{alg:hifm_loss}
\small
\begin{algorithmic}[1]
\Require Training targets with cached Hessian eigensystems;
         flow-matching model $\mathbf{v}_\theta$;
         time distribution $p(t)$;
         $c\geq1$, $\kappa>0$.

\State Compute positive rates $\alpha_i$ from each target's
       Hessian eigenvalues using $c$
       (\cref{app:homotopy_variants}).
\State For each target, set
       $\alpha_{\min}\gets\min_i\alpha_i$ and
       $r_i\gets\alpha_i/(\kappa\alpha_{\min})$.

\For{each training iteration}
    \State Sample a target $\mathbf{x}_1$ with its Hessian
           eigenvectors $\mathbf{P}$ and rates $r_i$.
    \State Sample $\mathbf{x}_0\sim p_0$ and $t\sim p(t)$.
    \State $\displaystyle
           \mathbf{x}_t\gets\mathbf{x}_1+
           \mathbf{P}\operatorname{diag}\!\left((1-t)^{r_i}\right)
           \mathbf{P}^{\top}(\mathbf{x}_0-\mathbf{x}_1)$.
    \State $\displaystyle
           \mathbf{u}_t\gets-
           \mathbf{P}\operatorname{diag}\!\left(
           r_i(1-t)^{r_i-1}\right)
           \mathbf{P}^{\top}(\mathbf{x}_0-\mathbf{x}_1)$.
    \State Take an optimizer step using $\mathbf{x}_t$, $t$, and
           $\mathbf{u}_t$ as conditional flow-matching supervision.
\EndFor
\State \Return $\theta$.
\end{algorithmic}
\end{algorithm}

\subsection{Sampling and Statistics}
\label{app:sampling_statistics}

We evaluate the final EMA checkpoint of each run on the test sets, without validation-based checkpoint selection. For each trained model, we generate three independent sets of
$10{,}000$ molecules using different seeds.
Generation is conditioned only on molecular size as SemlaFlow~\citep{irwin2024semlaflow}. For Hessian-QM9,
atom counts are sampled with replacement from the test set, following
the size-conditioned evaluation setup of SemlaFlow~\citep{irwin2024semlaflow}.
For GEOM-Drugs, each replicate uses $10{,}000$ distinct test molecular
identities, with one reference conformation selected per identity.
The atom count of each reference conformation defines the corresponding
generation size. For a given sampling seed, the same reference molecules
are used across all methods and training-data fractions.

Coordinates are generated with $100$ Euler steps using
\begin{equation}
    t_i = 1 - 100^{-i/100},
    \qquad i=0,\ldots,100,
\end{equation}
which places more integration steps near the data endpoint. We return
the final endpoint prediction as the generated structure. Categorical
variables follow the SemlaFlow sampling procedure, with noise level
and temperature set to one and self-conditioning enabled.

For Hessian-QM9, the three generation replicates are first averaged
within each trained model, and we report mean$\pm$sample standard
deviation across the three training runs. For GEOM-Drugs, where one
model is trained per setting, we report mean$\pm$population standard
deviation across the three generation replicates.

\subsection{Conditional Path Implementation}
\label{app:homotopy_variants}

We describe the coordinate-path implementation used in the main
SemlaFlow experiments, including eigenvalue rescaling and
hyperbolization, coordinate handling, and homotopy variants.

For a fixed data sample $\mathbf{x}_1\sim q$, let $\mathbf{x}_0$
be a source sample from the Gaussian sampler shared by all methods.
With no added terminal noise in the linear path and $\beta_i=0$
for HI-FM, the coupled paths for $t\in[0,1)$ are
\begin{align}
    \mathbf{x}^{\mathrm{lin}}_t
    &=
    (1-t)\mathbf{x}_0+t\mathbf{x}_1,
    \\
    \mathbf{x}^{\mathrm H}_t
    &=
    \mathbf{x}_1+
    \mathbf{P}\operatorname{diag}\!\left[(1-t)^{r_i}\right]
    \mathbf{P}^{\top}(\mathbf{x}_0-\mathbf{x}_1).
\end{align}
Here, $\mathbf{P}$ contains the eigenvectors of the raw energy
Hessian $\mathbf{H}_1$ at $\mathbf{x}_1$, and
$r_i=\alpha_i/(\kappa\alpha_{\min})$ uses the processed rates
$\alpha_i>0$ defined below, with
$\alpha_{\min}=\min_i\alpha_i$.
Both paths use the same source sample, data conformation, and
coordinate frame.

\paragraph{Eigenvalue rescaling and hyperbolization.}
Let $\lambda_i$ denote the raw Hessian eigenvalues from
\cref{eq:target_hessian}. Negative eigenvalues are treated as
zero before active-mode selection. We select active modes using
the absolute threshold
$\mathcal{I}=\{i:\lambda_i>10^{-4}\}$,
requiring $\mathcal{I}\neq\emptyset$. Let
$m=\min_{i\in\mathcal{I}}\lambda_i$ and
$M=\max_{i\in\mathcal{I}}\lambda_i$.
We rescale the active eigenvalues using $c\geq1$:
\[
    \alpha_i
    =m+(c-1)m
      \frac{\lambda_i-m}{\max(M-m,10^{-12})},
    \qquad i\in\mathcal{I}.
\]
We then apply \emph{hyperbolization}: all inactive modes receive
the minimum rescaled active rate,
$\alpha_i=\alpha_{\min}$ for $i\notin\mathcal{I}$, where
$\alpha_{\min}=\min_{j\in\mathcal{I}}\alpha_j$.
The processed rates replace the raw eigenvalues in the HI-FM
formulas, and the corresponding matrix is reconstructed as
$\mathbf{P}\operatorname{diag}(\alpha_i)\mathbf{P}^{\top}$.
The eigenvectors are unchanged.

The main experiments use $c=2$, $\kappa=1$, and the finite-time
construction in \cref{sec:interpolant}, with the coordinate loss applied to all modes. With these values, $r_i=1+(\lambda_i-m)/(M-m)\in[1,2]$ for active modes
and $r_i=1$ for hyperbolized modes: rigid-body and the softest modes follow
the linear schedule $1-t$, while the stiffest modes contract as $(1-t)^2$. For numerical evaluation of $r_i$, the denominator
$\kappa\alpha_{\min}$ is replaced by
$\max(\kappa\alpha_{\min},10^{-12})$.
This guard leaves the exponents unchanged for the main settings.
We floor $1-t$ at $10^{-8}$ for numerical evaluation of the paths
and velocities.

\paragraph{Coordinate handling.}
Data conformations are randomly rotated and centered, and the Hessian eigenvectors are rotated with them; the eigenvalues are unchanged. Source sampling, centering, and padding otherwise follow the SemlaFlow implementation and are shared by all methods.

\paragraph{Homotopy construction.}
The energy Hessian describes the local energy landscape around a
data sample, while the source samples used to construct its
conditional path may lie far from that sample. We therefore
consider two ways of introducing the Hessian-informed path
gradually: along flow time or during training.

Let $\eta(t,e)\in[0,1]$ be the mixing coefficient, where $e$ is
the training epoch. Writing $\eta=\eta(t,e)$,
we define
\begin{equation}
    \mathbf{x}^{\eta}_t(e)
    =
    (1-\eta)\mathbf{x}^{\mathrm{lin}}_t
    +\eta\mathbf{x}^{\mathrm H}_t.
    \label{eq:homotopy_sample}
\end{equation}
The choices $\eta=0$ and $\eta=1$ recover the linear and
direct HI-FM paths, respectively. This construction interpolates
coupled samples rather than mixing their probability densities.

The conditional velocity target at $\mathbf{x}^{\eta}_t(e)$
is obtained by differentiating with respect to flow time:
\begin{equation}
\begin{aligned}
    \mathbf{u}^{\eta}_t(e)
    :=\partial_t\mathbf{x}^{\eta}_t(e)
    ={}&
    (1-\eta)\partial_t\mathbf{x}^{\mathrm{lin}}_t
    +\eta\partial_t\mathbf{x}^{\mathrm H}_t
    \\
    &+
    (\partial_t\eta)
    \bigl(\mathbf{x}^{\mathrm H}_t
          -\mathbf{x}^{\mathrm{lin}}_t\bigr).
\end{aligned}
\label{eq:homotopy_sample_velocity}
\end{equation}
The data sample $\mathbf{x}_1$, source sample $\mathbf{x}_0$,
and epoch $e$ are held fixed during differentiation.
Each component derivative is taken along its own sampling path;
in particular,
$\partial_t\mathbf{x}^{\mathrm H}_t
=\mathbf{u}^{\mathrm H}_t
(\mathbf{x}^{\mathrm H}_t\mid\mathbf{x}_1)$.
The resulting samples and velocities provide the inputs and
targets for conditional flow matching (\cref{eq:fm}).

\paragraph{Flow-time homotopy (\hifmt{}).}
We set $\eta(t,e)=a(t)=t$. The weight on the Hessian-informed
path increases toward the data endpoint, placing greater emphasis
on local curvature later in the transport.

\paragraph{Epoch homotopy (\hifme{}).}
We set $\eta(t,e)=b(e)$. Let $U_e$ be the number of completed
optimizer updates at the start of epoch $e$, and let
$B_{\mathrm h}>0$ denote the homotopy schedule horizon,
measured in optimizer updates. The coefficient follows a
quantized schedule:
\[
    b(e)=\frac{1}{20}
    \min\!\left(
        20,\left\lfloor
        20\min\!\left(\frac{U_e}{B_{\mathrm h}},1\right)
        \right\rfloor
    \right).
\]
It starts at zero and remains fixed throughout each epoch.
For Hessian-QM9, $B_{\mathrm h}=33{,}000$.
For GEOM-Drugs, $B_{\mathrm h}=522{,}800$ is retained during
continuation to $1{,}045{,}600$ updates, without restarting
the schedule. GEOM-Drugs reaches $b=1$ at the first epoch
starting at or after $B_{\mathrm h}$ completed updates and
retains it thereafter.
Since $b(e)$ is independent of flow time,
$\partial_t\eta=0$, and the velocity target is the weighted
sum of the two component path derivatives.

Direct HI-FM uses the full Hessian-informed path from its first
optimizer update, without a path warm start. Learning-rate
warm-up is separate from these path schedules and is reported
in \cref{tab:semla_model_training}.

For the analytical paths before numerical flooring,
$\mathbf{x}^{\eta}_0(e)=\mathbf{x}_0$ and
$\lim_{t\to1}\mathbf{x}^{\eta}_t(e)=\mathbf{x}_1$
for every epoch $e$ and any mixing coefficient in $[0,1]$.
Thus, both homotopies preserve the common source and data
endpoint while changing the intermediate paths.
The effect of these variants on learned-field geometry is
examined in \cref{app:hjac_alignment}.

\subsection{Integration with SemlaFlow}
\label{app:semla_integration}

We use SemlaFlow's coordinate-output parameterization and adapt
its regression target to the conditional path. Let
$f_\theta(\mathbf{x},t)$ denote the network's coordinate prediction,
with categorical and self-conditioning inputs suppressed.
For any conditional path considered above, write $\mathbf{x}_t$
for the path sample at $t\in[0,1)$ and
$\mathbf{u}_t=\partial_t\mathbf{x}_t$ for its velocity target.
We use the transformed coordinate target and learned sampling field
\begin{equation}
\begin{aligned}
    \widetilde{\mathbf{x}}_1(t)
    &= \mathbf{x}_t+(1-t)\mathbf{u}_t, \\
    \mathbf{v}_\theta(\mathbf{x},t)
    &= \frac{f_\theta(\mathbf{x},t)-\mathbf{x}}{1-t}.
\end{aligned}
\label{eq:app_endpoint_target}
\end{equation}
For the linear conditional path in
\cref{eq:linear_conditional_path},
$\widetilde{\mathbf{x}}_1(t)=\mathbf{x}_1$.
For HI-FM paths, it is generally a time-dependent regression
target rather than the original data conformation.

We train $f_\theta$ with SemlaFlow's masked mean-squared error between $f_\theta(\mathbf{x}_t,t)$ and $\widetilde{\mathbf{x}}_1(t)$. For each molecule, the coordinate and velocity residuals satisfy
\begin{equation}
    f_\theta(\mathbf{x}_t,t)-\widetilde{\mathbf{x}}_1(t)
    =(1-t)\bigl(\mathbf{v}_\theta(\mathbf{x}_t,t)-\mathbf{u}_t\bigr).
    \label{eq:app_endpoint_residual}
\end{equation}
Thus, the coordinate MSE is a velocity regression weighted by $(1-t)^2$. We sample $t\sim\mathrm{Beta}(2,1)$ without additional inverse-time
weighting, giving a time-integral weight of $2t(1-t)^2$.
The time distribution, loss reduction, categorical objectives,
and self-conditioning procedure are shared by SemlaFlow and all
HI-FM variants; only the coordinate path and its corresponding
regression target change.

\subsection{Evaluation Metrics}
\label{app:metrics}

\paragraph{Energy and strain.}
We use MMFF94 on sanitized predicted graphs with complete force-field
parameters, including formal charges and explicit hydrogens. Energy and
strain per atom are $E(\mathbf{x})/N$ and
$[E(\mathbf{x})-E(\mathbf{x}^{\star})]/N$, in
kcal\,mol$^{-1}$\,atom$^{-1}$. Relaxation uses $1{,}000$ iterations. We average per-molecule normalized scores: energy requires a finite
initial score, and strain additionally requires a finite relaxed score
and optimizer status zero. Strain is used without clipping. For context, reference samples drawn from the full training
cohorts have mean MMFF94 strain of $0.349$ and $0.363$
kcal\,mol$^{-1}$\,atom$^{-1}$ for Hessian-QM9 and GEOM-Drugs,
respectively. HI-FM reduces generated-sample strain relative
to SemlaFlow, while remaining above these training-data
reference levels. These comparisons concern MMFF94-based
geometric quality and do not establish improved
quantum-chemical accuracy.

\paragraph{Native graph metrics.}
Validity is the proportion of generated graphs that pass RDKit
sanitization. Atom stability measures the proportion of atoms satisfying
SemlaFlow's element- and formal-charge-dependent valence rules; molecule
stability requires every atom in a molecule to satisfy those rules.
Uniqueness is the proportion of distinct canonical SMILES among the
accepted SMILES. All graph metrics are reported as percentages.

\paragraph{Bond-length and bond-angle distances.}
We use MiDi-style weighted histogram distances~\citep{vignac2023midi}
on predicted graphs and coordinates, including explicit hydrogens.
This calculation includes graphs that fail RDKit sanitization.
Hessian-QM9 uses all $4{,}151$ test conformations as the reference
distribution; GEOM-Drugs uses the generation-seed-specific test
reference described above. For each
category, we sum absolute differences between normalized cumulative
histograms and multiply by the bin width.

Bond lengths use $0.01$\,\AA\ bins and single, double, triple, and
aromatic bond categories. Each category is weighted by its reference
bond count divided by the total number of unordered reference atom pairs,
including nonbonded pairs. Bond angles use $0.1^\circ$ bins and
central-element categories, weighted by
$p(z)[1-P(\nu=0\mid z)-P(\nu=1\mid z)]$ and normalized over
participating elements, where $\nu$ is bond-order valency. Tables display bond-length
$W_1$ in $10^{-3}$\,\AA\ and bond-angle $W_1$ in degrees.

\section{Additional Results}
\label{app:additional_results}
\label{app:qm9_semla_complete}
\label{app:geom_semla_complete}

\Cref{tab:semla_graph_quality} reports native atom stability, molecule
stability, and uniqueness on Hessian-QM9 and GEOM-Drugs. Atom stability
on both datasets and uniqueness on GEOM-Drugs are already close to
$100\%$, leaving limited room for further improvement. Across all
three metrics, HI-FM maintains high scores that are broadly comparable
to SemlaFlow throughout the data-fraction sweep.

\begin{table*}[h!]
    \centering
    \footnotesize
    \setlength{\tabcolsep}{3.5pt}
    \caption{Additional native graph metrics (\%) on the test sets, using final EMA weights.
    Values are mean$\pm$SD: sample SD across three training seeds for
    Hessian-QM9, and population SD across three generation replicates
    of one training seed for GEOM-Drugs. A $+$ prefix denotes an HI-FM
    variant of SemlaFlow.}
    \label{tab:semla_graph_quality}
    \label{tab:qm9_semla_extra}
    \label{tab:geom_semla_extra}
    \resizebox{\textwidth}{!}{%
    \begin{tabular}{clcccccc}
        \toprule
        & & \multicolumn{3}{c}{Hessian-QM9}
        & \multicolumn{3}{c}{GEOM-Drugs} \\
        \cmidrule(lr){3-5}
        \cmidrule(lr){6-8}
        Data & Method
        & \shortstack{Atom\\stability $\uparrow$}
        & \shortstack{Molecule\\stability $\uparrow$}
        & Uniqueness $\uparrow$
        & \shortstack{Atom\\stability $\uparrow$}
        & \shortstack{Molecule\\stability $\uparrow$}
        & Uniqueness $\uparrow$ \\
        \midrule

        $10\%$ & SemlaFlow
        & $99.157 \pm 0.189$ & $96.90 \pm 0.25$ & $96.049 \pm 0.245$
        & $99.894 \pm 0.005$ & $95.71 \pm 0.12$ & $99.941 \pm 0.023$ \\

        & $+$\hifmd{}
        & $99.247 \pm 0.246$ & $96.95 \pm 0.18$ & $96.178 \pm 0.081$
        & $99.876 \pm 0.007$ & $94.99 \pm 0.25$ & $99.971 \pm 0.014$ \\

        & $+$\hifmt{}
        & $99.151 \pm 0.213$ & $96.69 \pm 0.12$ & $96.176 \pm 0.177$
        & $99.883 \pm 0.005$ & $95.18 \pm 0.13$ & $99.967 \pm 0.009$ \\

        & $+$\hifme{}
        & $99.212 \pm 0.219$ & $96.90 \pm 0.10$ & $96.135 \pm 0.219$
        & $99.885 \pm 0.004$ & $95.37 \pm 0.10$ & $99.963 \pm 0.023$ \\
        \midrule

        $25\%$ & SemlaFlow
        & $99.159 \pm 0.191$ & $96.98 \pm 0.08$ & $96.422 \pm 0.044$
        & $99.927 \pm 0.004$ & $96.98 \pm 0.18$ & $99.982 \pm 0.005$ \\

        & $+$\hifmd{}
        & $99.190 \pm 0.228$ & $96.75 \pm 0.22$ & $96.405 \pm 0.084$
        & $99.906 \pm 0.003$ & $96.19 \pm 0.09$ & $99.982 \pm 0.005$ \\

        & $+$\hifmt{}
        & $99.177 \pm 0.174$ & $96.84 \pm 0.12$ & $96.465 \pm 0.034$
        & $99.919 \pm 0.006$ & $96.63 \pm 0.20$ & $99.993 \pm 0.005$ \\

        & $+$\hifme{}
        & $99.156 \pm 0.182$ & $96.80 \pm 0.13$ & $96.360 \pm 0.163$
        & $99.907 \pm 0.004$ & $96.34 \pm 0.21$ & $99.982 \pm 0.014$ \\
        \midrule

        $50\%$ & SemlaFlow
        & $99.904 \pm 0.014$ & $98.70 \pm 0.14$ & $96.337 \pm 0.031$
        & $99.935 \pm 0.001$ & $97.31 \pm 0.05$ & $99.989 \pm 0.000$ \\

        & $+$\hifmd{}
        & $99.888 \pm 0.017$ & $98.28 \pm 0.26$ & $96.516 \pm 0.097$
        & $99.933 \pm 0.003$ & $97.32 \pm 0.18$ & $99.993 \pm 0.005$ \\

        & $+$\hifmt{}
        & $99.897 \pm 0.008$ & $98.48 \pm 0.06$ & $96.384 \pm 0.060$
        & $99.934 \pm 0.003$ & $97.26 \pm 0.14$ & $99.982 \pm 0.018$ \\

        & $+$\hifme{}
        & $99.908 \pm 0.010$ & $98.61 \pm 0.15$ & $96.431 \pm 0.153$
        & $99.932 \pm 0.003$ & $97.13 \pm 0.11$ & $99.989 \pm 0.009$ \\
        \midrule

        $100\%$ & SemlaFlow
        & $99.945 \pm 0.009$ & $99.05 \pm 0.19$ & $96.821 \pm 0.103$
        & $99.952 \pm 0.003$ & $98.06 \pm 0.11$ & $99.996 \pm 0.005$ \\

        & $+$\hifmd{}
        & $99.930 \pm 0.011$ & $98.77 \pm 0.19$ & $96.997 \pm 0.079$
        & $99.934 \pm 0.003$ & $97.30 \pm 0.07$ & $99.989 \pm 0.009$ \\

        & $+$\hifmt{}
        & $99.931 \pm 0.007$ & $98.78 \pm 0.14$ & $96.869 \pm 0.196$
        & $99.937 \pm 0.006$ & $97.44 \pm 0.18$ & $99.993 \pm 0.010$ \\

        & $+$\hifme{}
        & $99.944 \pm 0.006$ & $99.01 \pm 0.09$ & $96.889 \pm 0.193$
        & $99.945 \pm 0.002$ & $97.71 \pm 0.11$ & $99.993 \pm 0.005$ \\
        \bottomrule
    \end{tabular}%
    }
\end{table*}

\subsection{Additional Backbone: EDM-Style EGNN}
\label{app:qm9_egnn}

To examine whether the benefits of HI-FM extend beyond SemlaFlow,
we evaluate an EDM-style equivariant graph neural network (EGNN)~\citep{Satorras2021EGNN,Hoogeboom2022EDM}
on Hessian-QM9. All settings use flow matching training rather than
the original EDM diffusion objective.

For the Hessian-QM9 EGNN experiments, OT-FM and all three HI-FM
variants use the same nine-layer EGNN with hidden width 256.
Training uses Adam with learning-rate warm-up from $10^{-5}$
to $1.5\times10^{-4}$, followed by cosine decay toward
$3\times10^{-5}$. Architecture and optimizer settings are shared
across methods and data fractions. Budgets are identical across methods
within each fraction and approximately matched across fractions.

We use three training seeds for each conditional path and data
fraction. Each trained model generates $10{,}000$ molecules using
Euler integration with $100$ steps. Here we report per-run median energy and strain, and compute the chemical metrics with the EDM evaluation code.

\Cref{tab:qm9_egnn} shows consistent gains with this second
backbone. At every data fraction, all three HI-FM paths reduce
seed-averaged median energy and strain per atom and improve mean
validity, atom stability, and molecule stability relative to
OT-FM. Across the twelve matched-data comparisons, energy is
reduced by $4.9$-$12.7\%$ and strain by $10.1$-$18.2\%$.
These results show that the benefits of HI-FM also occur with
a second equivariant backbone.

\begin{table*}[t!]
    \centering
    \footnotesize
    \setlength{\tabcolsep}{4pt}
    \caption{
        \textbf{HI-FM with an EDM-style EGNN on Hessian-QM9.}
        Values are mean$\pm$sample standard deviation across three
        training seeds, with $10{,}000$ generated molecules per seed.
        Energy and strain are per-run medians in
        kcal\,mol$^{-1}$\,atom$^{-1}$; chemical metrics are percentages.
        Bold indicates the best mean within each data fraction.
    }
    \label{tab:qm9_egnn}
    \resizebox{\textwidth}{!}{%
    \begin{tabular}{clccccc}
        \toprule
        Data & Conditional path
        & Energy/atom $\downarrow$
        & Strain/atom $\downarrow$
        & Validity (\%) $\uparrow$
        & Atom stability (\%) $\uparrow$
        & Molecule stability (\%) $\uparrow$ \\
        \midrule

        $10\%$ & OT-FM
        & $2.6100 \pm 0.0805$
        & $1.2333 \pm 0.0747$
        & $68.05 \pm 2.32$
        & $89.73 \pm 0.74$
        & $24.74 \pm 1.58$ \\

        $10\%$ & \hifmd{}
        & $\mathbf{2.2792 \pm 0.0368}$
        & $\mathbf{1.0091 \pm 0.0519}$
        & $71.23 \pm 3.47$
        & $\mathbf{91.33 \pm 0.36}$
        & $\mathbf{28.26 \pm 1.97}$ \\

        $10\%$ & \hifmt{}
        & $2.3580 \pm 0.0112$
        & $1.0790 \pm 0.0330$
        & $\mathbf{71.77 \pm 2.84}$
        & $91.21 \pm 0.47$
        & $27.89 \pm 1.59$ \\

        $10\%$ & \hifme{}
        & $2.2991 \pm 0.0595$
        & $1.0411 \pm 0.0608$
        & $71.74 \pm 1.86$
        & $91.21 \pm 0.30$
        & $26.84 \pm 1.59$ \\
        \midrule

        $25\%$ & OT-FM
        & $2.3694 \pm 0.0391$
        & $1.0569 \pm 0.0208$
        & $74.92 \pm 2.22$
        & $92.40 \pm 0.03$
        & $28.87 \pm 0.40$ \\

        $25\%$ & \hifmd{}
        & $2.1475 \pm 0.0792$
        & $0.8709 \pm 0.0313$
        & $76.37 \pm 1.12$
        & $\mathbf{93.37 \pm 0.15}$
        & $\mathbf{33.51 \pm 0.46}$ \\

        $25\%$ & \hifmt{}
        & $2.2536 \pm 0.0982$
        & $0.9505 \pm 0.0219$
        & $\mathbf{76.86 \pm 1.96}$
        & $93.31 \pm 0.19$
        & $32.73 \pm 1.32$ \\

        $25\%$ & \hifme{}
        & $\mathbf{2.1283 \pm 0.0901}$
        & $\mathbf{0.8641 \pm 0.0263}$
        & $76.72 \pm 1.34$
        & $93.30 \pm 0.47$
        & $32.98 \pm 1.71$ \\
        \midrule

        $50\%$ & OT-FM
        & $2.2801 \pm 0.1225$
        & $0.9799 \pm 0.0451$
        & $76.47 \pm 2.18$
        & $93.28 \pm 0.08$
        & $32.51 \pm 0.69$ \\

        $50\%$ & \hifmd{}
        & $2.0875 \pm 0.0474$
        & $\mathbf{0.8015 \pm 0.0144}$
        & $77.99 \pm 1.84$
        & $\mathbf{94.17 \pm 0.14}$
        & $36.32 \pm 0.41$ \\

        $50\%$ & \hifmt{}
        & $2.1649 \pm 0.1301$
        & $0.8588 \pm 0.0625$
        & $78.30 \pm 1.29$
        & $94.14 \pm 0.21$
        & $36.39 \pm 1.74$ \\

        $50\%$ & \hifme{}
        & $\mathbf{2.0511 \pm 0.0639}$
        & $0.8168 \pm 0.0383$
        & $\mathbf{79.20 \pm 0.52}$
        & $94.11 \pm 0.29$
        & $\mathbf{36.47 \pm 1.45}$ \\
        \midrule

        $100\%$ & OT-FM
        & $2.2406 \pm 0.0714$
        & $0.9428 \pm 0.0238$
        & $77.35 \pm 1.81$
        & $93.53 \pm 0.35$
        & $33.83 \pm 1.44$ \\

        $100\%$ & \hifmd{}
        & $\mathbf{2.0372 \pm 0.0971}$
        & $\mathbf{0.7923 \pm 0.0548}$
        & $\mathbf{80.02 \pm 1.55}$
        & $94.10 \pm 0.41$
        & $36.45 \pm 1.81$ \\

        $100\%$ & \hifmt{}
        & $2.0922 \pm 0.0799$
        & $0.8179 \pm 0.0538$
        & $79.98 \pm 0.29$
        & $\mathbf{94.39 \pm 0.12}$
        & $\mathbf{38.25 \pm 2.35}$ \\

        $100\%$ & \hifme{}
        & $2.0597 \pm 0.0581$
        & $0.8106 \pm 0.0315$
        & $79.56 \pm 1.88$
        & $94.31 \pm 0.33$
        & $37.11 \pm 1.10$ \\
        \bottomrule
    \end{tabular}%
    }
\end{table*}

\subsection{Hessian-Jacobian Alignment}
\label{app:hjac_alignment}

We examine whether the learned VF reflects the target
energy curvature near the data endpoint ($t\approx1$). We compare
its local Jacobian with the target Hessian, measuring both their
dominant directions and their overall alignment.

\paragraph{Protocol.}
For each dataset, we use the same $500$ validation molecules across
all methods and training-data fractions, with one fixed conformation
per GEOM-Drugs molecule. We evaluate final EMA models at $t\approx1$, using two probe conditions: \emph{clean} probes at the target
conformations and \emph{common linear} probes constructed as
$\mathbf{x}_t=(1-t)\mathbf{x}_0+t\mathbf{x}_1$ from a fixed Gaussian
source and target pair. The same probe coordinates are used for all
methods, allowing their learned Jacobians to be compared at identical
input locations.

Let $\mathbf{C}$ denote coordinate centering and $\mathbf{z}$ the
categorical inputs. With self-conditioning set to zero and $\mathbf{z}$ held fixed,
we differentiate the endpoint-derived field
\begin{equation}
   \mathbf{v}_{\theta}(\mathbf{x},t;\mathbf{z})
    =
    \frac{
    f_{\theta,\mathrm{EMA}}(\mathbf{C}\mathbf{x},t;\mathbf{z})
    -\mathbf{C}\mathbf{x}
    }{1-t}.
    \qquad
    J=\frac{\partial\mathbf{v}_{\theta}}{\partial\mathbf{x}}.
    \label{eq:hjac_field}
\end{equation}

\paragraph{Alignment metrics.}
After removing padded coordinates, we symmetrize the Jacobian and
Hessian as $J_s=(J+J^\top)/2$ and $H_s=(H+H^\top)/2$.
Let $W$ contain the eigenvectors of $H_s$ with eigenvalues exceeding
$\max(10^{-6},10^{-4}\lambda_{\max}(H_s))$, and define
\begin{equation}
    A=-W^\top J_sW, \qquad B=W^\top H_sW.
    \label{eq:hjac_projection}
\end{equation}
We compute
\begin{align}
    \mathrm{ProjSim@10}
    &=\frac{\lVert U_B^\top U_A\rVert_F^2}{10},
    \label{eq:hjac_projsim}\\
    \mathrm{HessianFit}
    &=\frac{\langle A,B\rangle_F}
    {\lVert A\rVert_F\lVert B\rVert_F+10^{-12}},
    \label{eq:hjac_fit}
\end{align}
where $U_A$ and $U_B$ contain the eigenvectors corresponding to the
ten largest algebraic eigenvalues of $A$ and $B$, respectively.
ProjSim@10 measures overlap between the dominant ten-dimensional
subspaces, while HessianFit measures Frobenius-cosine alignment over
the retained positive-curvature subspace. Higher values indicate
closer alignment for both metrics.

\paragraph{Results.}
\Cref{tab:qm9_hjac,tab:geom_hjac} report the complete results
across training-data fractions and probe conditions.
The full-data shared-linear results are summarized in
\cref{tab:hjac_main}.
All three HI-FM variants achieve higher mean alignment on both
metrics at every training-data fraction under both probe
conditions. Flow-time homotopy achieves the highest mean
of both metrics in every shared-linear comparison.

\begin{table*}[t]
    \centering
    \footnotesize
    \setlength{\tabcolsep}{4pt}
    \caption{Hessian-Jacobian alignment on Hessian-QM9 at $t\approx1$,
    using final EMA weights. Values are mean$\pm$sample SD across
    three training-seed means, each computed over $500$ validation
    molecules. Bold marks the best mean within each fraction and probe.}
    \label{tab:qm9_hjac}
    \resizebox{\textwidth}{!}{%
    \begin{tabular}{clcccc}
        \toprule
        & & \multicolumn{2}{c}{Clean probes}
        & \multicolumn{2}{c}{Shared-linear probes} \\
        \cmidrule(lr){3-4}\cmidrule(lr){5-6}
        Data & Method
        & ProjSim@10 $\uparrow$ & HessianFit $\uparrow$
        & ProjSim@10 $\uparrow$ & HessianFit $\uparrow$ \\
        \midrule
        $10\%$ & SemlaFlow
        & $0.3107 \pm 0.0102$ & $0.6863 \pm 0.0016$
        & $0.3340 \pm 0.0069$ & $0.7049 \pm 0.0010$ \\

         & SemlaFlow-HI-FM-D
        & $0.7427 \pm 0.0112$ & $0.7513 \pm 0.0011$
        & $0.7612 \pm 0.0090$ & $0.7693 \pm 0.0028$ \\

         & SemlaFlow-HI-FM-T
        & $\mathbf{0.7731 \pm 0.0069}$ & $\mathbf{0.7645 \pm 0.0027}$
        & $\mathbf{0.7924 \pm 0.0078}$ & $\mathbf{0.7819 \pm 0.0017}$ \\

         & SemlaFlow-HI-FM-E
        & $0.7139 \pm 0.0158$ & $0.7469 \pm 0.0026$
        & $0.7334 \pm 0.0051$ & $0.7621 \pm 0.0009$ \\

        \midrule
        $25\%$ & SemlaFlow
        & $0.3119 \pm 0.0244$ & $0.6895 \pm 0.0009$
        & $0.3319 \pm 0.0323$ & $0.7045 \pm 0.0015$ \\

         & SemlaFlow-HI-FM-D
        & $0.7416 \pm 0.0229$ & $0.7544 \pm 0.0008$
        & $0.7593 \pm 0.0203$ & $0.7702 \pm 0.0019$ \\

         & SemlaFlow-HI-FM-T
        & $\mathbf{0.7727 \pm 0.0062}$ & $\mathbf{0.7648 \pm 0.0022}$
        & $\mathbf{0.7881 \pm 0.0049}$ & $\mathbf{0.7805 \pm 0.0005}$ \\

         & SemlaFlow-HI-FM-E
        & $0.7202 \pm 0.0050$ & $0.7422 \pm 0.0033$
        & $0.7295 \pm 0.0116$ & $0.7590 \pm 0.0026$ \\

        \midrule
        $50\%$ & SemlaFlow
        & $0.3313 \pm 0.0149$ & $0.6911 \pm 0.0042$
        & $0.3568 \pm 0.0086$ & $0.7069 \pm 0.0011$ \\

         & SemlaFlow-HI-FM-D
        & $0.7635 \pm 0.0078$ & $0.7559 \pm 0.0019$
        & $0.7786 \pm 0.0078$ & $0.7710 \pm 0.0016$ \\

         & SemlaFlow-HI-FM-T
        & $\mathbf{0.7809 \pm 0.0063}$ & $\mathbf{0.7648 \pm 0.0077}$
        & $\mathbf{0.8002 \pm 0.0053}$ & $\mathbf{0.7824 \pm 0.0061}$ \\

         & SemlaFlow-HI-FM-E
        & $0.7078 \pm 0.0180$ & $0.7439 \pm 0.0062$
        & $0.7268 \pm 0.0176$ & $0.7615 \pm 0.0064$ \\

        \midrule
        $100\%$ & SemlaFlow
        & $0.2884 \pm 0.0147$ & $0.6884 \pm 0.0026$
        & $0.3155 \pm 0.0118$ & $0.7046 \pm 0.0014$ \\

         & SemlaFlow-HI-FM-D
        & $0.7716 \pm 0.0045$ & $0.7564 \pm 0.0051$
        & $0.7871 \pm 0.0104$ & $0.7716 \pm 0.0021$ \\

         & SemlaFlow-HI-FM-T
        & $\mathbf{0.7850 \pm 0.0101}$ & $\mathbf{0.7689 \pm 0.0029}$
        & $\mathbf{0.7972 \pm 0.0042}$ & $\mathbf{0.7846 \pm 0.0012}$ \\

         & SemlaFlow-HI-FM-E
        & $0.7200 \pm 0.0280$ & $0.7458 \pm 0.0019$
        & $0.7436 \pm 0.0207$ & $0.7622 \pm 0.0039$ \\

        \bottomrule
    \end{tabular}%
    }
\end{table*}

\begin{table*}[t]
    \centering
    \footnotesize
    \setlength{\tabcolsep}{4pt}
    \caption{Hessian-Jacobian alignment on GEOM-Drugs at $t\approx1$,
    using final EMA weights. Values are mean$\pm$population SD across
    $500$ validation molecules for one training seed. Bold marks the
    best mean within each fraction and probe.}
    \label{tab:geom_hjac}
    \resizebox{\textwidth}{!}{%
    \begin{tabular}{clcccc}
        \toprule
        & & \multicolumn{2}{c}{Clean probes}
        & \multicolumn{2}{c}{Shared-linear probes} \\
        \cmidrule(lr){3-4}\cmidrule(lr){5-6}
        Data & Method
        & ProjSim@10 $\uparrow$ & HessianFit $\uparrow$
        & ProjSim@10 $\uparrow$ & HessianFit $\uparrow$ \\
        \midrule
        $10\%$ & SemlaFlow
        & $0.1166 \pm 0.0493$ & $0.6838 \pm 0.0222$
        & $0.1465 \pm 0.0567$ & $0.6901 \pm 0.0111$ \\

         & SemlaFlow-HI-FM-D
        & $0.5442 \pm 0.1105$ & $0.7525 \pm 0.0351$
        & $0.5721 \pm 0.0981$ & $0.7581 \pm 0.0107$ \\

         & SemlaFlow-HI-FM-T
        & $\mathbf{0.5825 \pm 0.1101}$ & $\mathbf{0.7605 \pm 0.0308}$
        & $\mathbf{0.6308 \pm 0.0952}$ & $\mathbf{0.7687 \pm 0.0108}$ \\

         & SemlaFlow-HI-FM-E
        & $0.5031 \pm 0.1078$ & $0.7537 \pm 0.0253$
        & $0.5477 \pm 0.1019$ & $0.7588 \pm 0.0109$ \\

        \midrule
        $25\%$ & SemlaFlow
        & $0.1056 \pm 0.0475$ & $0.6863 \pm 0.0220$
        & $0.1211 \pm 0.0563$ & $0.6890 \pm 0.0109$ \\

         & SemlaFlow-HI-FM-D
        & $0.5190 \pm 0.1160$ & $0.7382 \pm 0.0596$
        & $0.6041 \pm 0.1010$ & $0.7557 \pm 0.0136$ \\

         & SemlaFlow-HI-FM-T
        & $\mathbf{0.5236 \pm 0.1128}$ & $\mathbf{0.7627 \pm 0.0368}$
        & $\mathbf{0.6239 \pm 0.0929}$ & $\mathbf{0.7716 \pm 0.0116}$ \\

         & SemlaFlow-HI-FM-E
        & $0.5086 \pm 0.0969$ & $0.7523 \pm 0.0318$
        & $0.5868 \pm 0.0907$ & $0.7611 \pm 0.0111$ \\

        \midrule
        $50\%$ & SemlaFlow
        & $0.1096 \pm 0.0571$ & $0.6879 \pm 0.0275$
        & $0.1201 \pm 0.0595$ & $0.6893 \pm 0.0108$ \\

         & SemlaFlow-HI-FM-D
        & $0.5283 \pm 0.1017$ & $0.7551 \pm 0.0394$
        & $0.5893 \pm 0.0981$ & $0.7625 \pm 0.0118$ \\

         & SemlaFlow-HI-FM-T
        & $\mathbf{0.5598 \pm 0.1014}$ & $\mathbf{0.7641 \pm 0.0514}$
        & $\mathbf{0.6211 \pm 0.0841}$ & $\mathbf{0.7750 \pm 0.0103}$ \\

         & SemlaFlow-HI-FM-E
        & $0.4520 \pm 0.1047$ & $0.7415 \pm 0.0673$
        & $0.5108 \pm 0.0970$ & $0.7552 \pm 0.0141$ \\

        \midrule
        $100\%$ & SemlaFlow
        & $0.0777 \pm 0.0415$ & $0.6820 \pm 0.0341$
        & $0.1148 \pm 0.0590$ & $0.6893 \pm 0.0110$ \\

         & SemlaFlow-HI-FM-D
        & $0.5428 \pm 0.1259$ & $0.7502 \pm 0.0494$
        & $0.6033 \pm 0.0995$ & $0.7621 \pm 0.0117$ \\

         & SemlaFlow-HI-FM-T
        & $0.5420 \pm 0.1075$ & $\mathbf{0.7668 \pm 0.0259}$
        & $\mathbf{0.6473 \pm 0.0882}$ & $\mathbf{0.7767 \pm 0.0104}$ \\

         & SemlaFlow-HI-FM-E
        & $\mathbf{0.5756 \pm 0.1084}$ & $0.7588 \pm 0.0308$
        & $0.6294 \pm 0.0964$ & $0.7671 \pm 0.0121$ \\

        \bottomrule
    \end{tabular}%
    }
\end{table*}

\section{Ablation Studies}
\label{app:ablations}

We evaluate computational overhead, the role of Hessian eigendirections,
and sensitivity to the path parameters using direct HI-FM on the
$10\%$ Hessian-QM9 training subset. Generation-quality ablations use
the validation set for molecular-size sampling and reference
distributions. The remaining settings follow
\cref{app:model_settings,app:sampling_statistics,app:metrics}, with
$33{,}000$ optimizer updates and three training seeds. Each final EMA
model produces three replicates of $10{,}000$ molecules. We average
the generation replicates within each model and report mean$\pm$sample
standard deviation across training seeds.

\begin{figure}[t]
    \centering
    \includegraphics[width=0.75\linewidth]
        {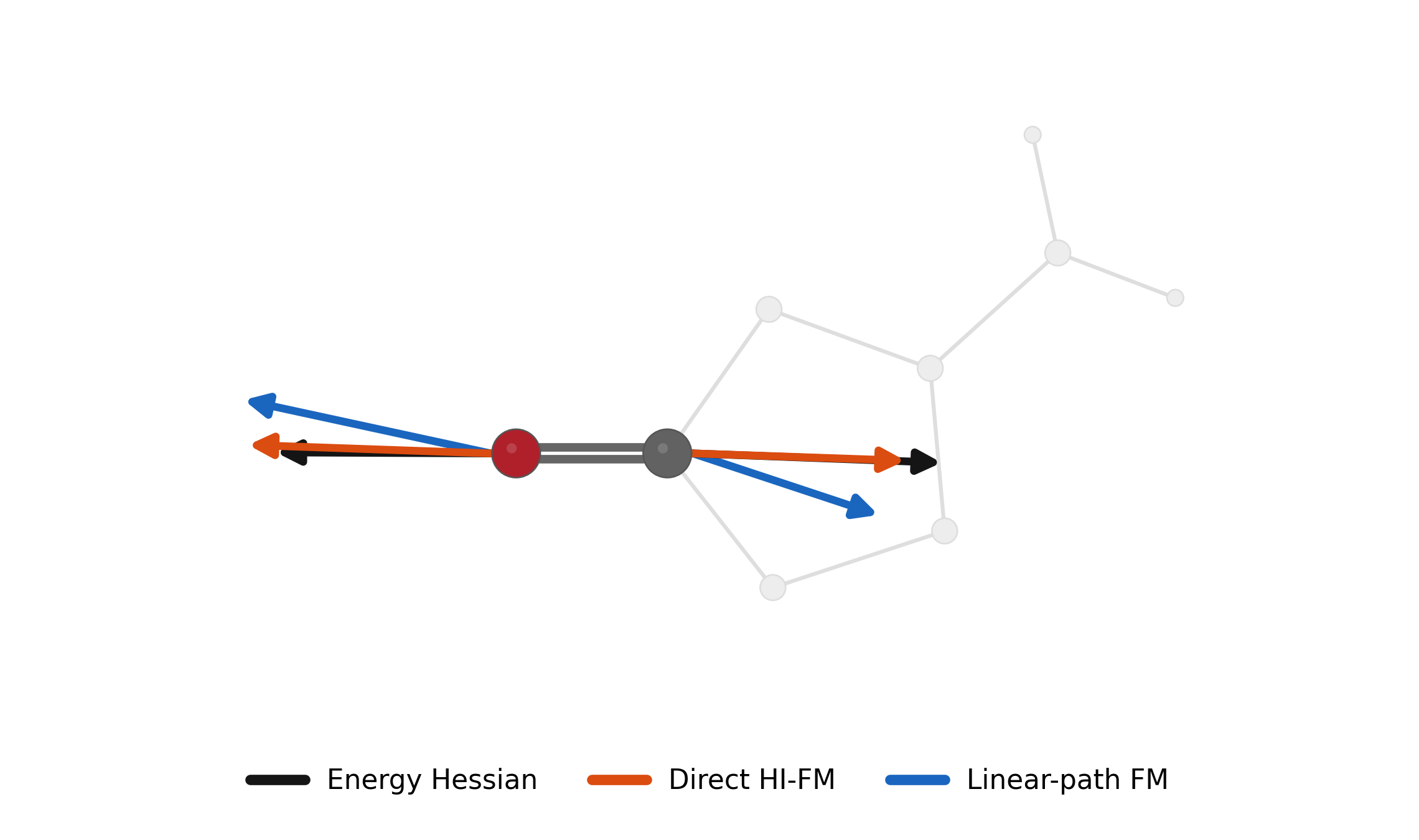}
    \caption{
        \textbf{A molecular example of Hessian-mode alignment.}
        For one Hessian-QM9 conformation, we compare a
        C=O-stretching energy-Hessian mode (black) with learned
        deformation directions from direct HI-FM (orange) and
        linear-path SemlaFlow (blue).
        Arrows show the components on the highlighted carbon and oxygen;
        the remaining molecular structure is shown in gray for context.
        In this example, HI-FM more closely follows the
        Hessian C=O-stretching direction.
        Arrows represent deformation modes, not generation trajectories.
    }
    \label{fig:co_stretch_alignment}
\end{figure}

\subsection{Computational Overhead}
\label{app:computational_overhead}

HI-FM uses precomputed target energy Hessians. On Hessian-QM9, their
eigendecompositions are also cached before training and reused to
construct the conditional paths and targets. Direct HI-FM incurs a
mean paired overhead of $18.9\%$ per optimizer update, with timings
reported in \cref{tab:hifm_compute}. Generation requires no energy
or Hessian evaluations.

We time SemlaFlow and direct HI-FM on the same GPU for three paired seeds, with two runs per method and seed after a short warm-up. The timings cover the complete training update, including data loading and path construction, and exclude dataset preparation, validation, logging, and checkpointing.

\begin{table}[h]
    \centering
    \footnotesize
    \caption{Training time on $10\%$ of Hessian-QM9 with cached
    eigensystems. Mean$\pm$sample standard deviation over three seeds,
    after averaging two timing repetitions per seed.}
    \label{tab:hifm_compute}
    \begin{tabular}{lc}
        \toprule
        Method & Time per optimizer update (s) \\
        \midrule
        SemlaFlow
        & $0.3062 \pm 0.0015$ \\

        SemlaFlow-\hifmd{}
        & $0.3640 \pm 0.0034$ \\
        \bottomrule
    \end{tabular}
\end{table}

\subsection{Role of Hessian Eigendirections}
\label{app:directional_control}

We test whether the target eigendirections contribute beyond the
Hessian spectrum. For each conformation, the eigenvectors $U_+$ with
eigenvalues above $10^{-4}$ are replaced by $U_+Q$, where $Q$ is a
Haar-distributed orthogonal matrix. Eigenvalues and the remaining
eigenvectors are unchanged. This preserves the spectrum and the
active and low-mode subspaces while randomizing the directions
associated with the active eigenvalues. The random basis is sampled
once per molecule and training seed and follows the usual spatial
augmentation. Both HI-FM settings use $c=2$ and $\kappa=1$.

\begin{table*}[h!]
    \centering
    \footnotesize
    \setlength{\tabcolsep}{4pt}
    \caption{Effect of randomizing Hessian eigendirections on $10\%$
    of Hessian-QM9. Eigenvalues are unchanged. Bold indicates the
    best mean in each column.}
    \label{tab:hifm_directions}
    \resizebox{\textwidth}{!}{%
    \begin{tabular}{lccccc}
        \toprule
        Method
        & Energy/atom $\downarrow$
        & Strain/atom $\downarrow$
        & Validity (\%) $\uparrow$
        & \shortstack{Bond-length $W_1$\\($10^{-3}$ \AA) $\downarrow$}
        & \shortstack{Bond-angle $W_1$\\($^\circ$) $\downarrow$} \\
        \midrule

        SemlaFlow
        & $2.348 \pm 0.033$
        & $1.266 \pm 0.009$
        & $94.57 \pm 0.59$
        & $1.68 \pm 0.14$
        & $1.137 \pm 0.082$ \\

        $+$\hifmd{}
        & $\mathbf{2.103 \pm 0.025}$
        & $\mathbf{0.996 \pm 0.012}$
        & $\mathbf{95.30 \pm 0.39}$
        & $\mathbf{1.23 \pm 0.07}$
        & $\mathbf{0.930 \pm 0.128}$ \\

        $+$\hifmd{} (randomized directions)
        & $2.289 \pm 0.073$
        & $1.191 \pm 0.054$
        & $94.68 \pm 0.34$
        & $1.47 \pm 0.04$
        & $1.078 \pm 0.185$ \\

        \bottomrule
    \end{tabular}%
    }
\end{table*}

\Cref{tab:hifm_directions} shows that the original eigendirections
achieve the best mean on all five metrics. Randomizing them removes
approximately $76\%$ of the energy gain and $72\%$ of the strain
gain over SemlaFlow. The randomized control retains smaller
improvements, suggesting that the target-specific eigendirections
provide useful information beyond the retained spectrum and subspaces.

\subsection{Hyperparameter Sensitivity}
\label{app:hyperparameter_sensitivity}

We vary $c$ and $\kappa$ individually around the main-paper default
$(c,\kappa)=(2,1)$, keeping the original target eigensystems and all
other training and sampling settings fixed. The parameter $c$ controls condition-number rescaling, while $\kappa$ rescales the mode exponents through the clock in \cref{eq:hifm_clock}.

\begin{table*}[h!]
    \centering
    \footnotesize
    \setlength{\tabcolsep}{4pt}
    \caption{Sensitivity to $c$ and $\kappa$ on $10\%$ of
    Hessian-QM9. Parameters are varied individually around the
    main-paper default $(2,1)$. Bold indicates the best mean.}
    \label{tab:hifm_sensitivity}
    \resizebox{\textwidth}{!}{%
    \begin{tabular}{ccccccc}
        \toprule
        $c$ & $\kappa$
        & Energy/atom $\downarrow$
        & Strain/atom $\downarrow$
        & Validity (\%) $\uparrow$
        & \shortstack{Bond-length $W_1$\\($10^{-3}$ \AA) $\downarrow$}
        & \shortstack{Bond-angle $W_1$\\($^\circ$) $\downarrow$} \\
        \midrule

        \multicolumn{2}{c}{SemlaFlow}
        & $2.348 \pm 0.033$
        & $1.266 \pm 0.009$
        & $94.57 \pm 0.59$
        & $1.68 \pm 0.14$
        & $1.137 \pm 0.082$ \\
        \midrule

        $2$ & $1$
        & $\mathbf{2.103 \pm 0.025}$
        & $\mathbf{0.996 \pm 0.012}$
        & $\mathbf{95.30 \pm 0.39}$
        & $\mathbf{1.23 \pm 0.07}$
        & $\mathbf{0.930 \pm 0.128}$ \\

        $1.5$ & $1$
        & $2.173 \pm 0.047$
        & $1.040 \pm 0.044$
        & $94.97 \pm 0.12$
        & $1.27 \pm 0.11$
        & $1.008 \pm 0.125$ \\

        $5$ & $1$
        & $2.705 \pm 0.030$
        & $1.614 \pm 0.041$
        & $94.66 \pm 0.36$
        & $1.82 \pm 0.11$
        & $1.044 \pm 0.107$ \\

        $2$ & $0.5$
        & $2.630 \pm 0.041$
        & $1.499 \pm 0.171$
        & $92.73 \pm 0.87$
        & $2.08 \pm 0.09$
        & $1.214 \pm 0.202$ \\

        $2$ & $2$
        & $5.999 \pm 0.253$
        & $4.926 \pm 0.229$
        & $94.90 \pm 0.61$
        & $6.24 \pm 0.32$
        & $2.980 \pm 0.090$ \\

        \bottomrule
    \end{tabular}%
    }
\end{table*}

The default $(2,1)$ achieves the best mean on all five metrics among
the tested settings. With $c=1.5$, HI-FM continues to improve energy,
strain, and both geometry distances over SemlaFlow. Increasing $c$
to $5$ or changing $\kappa$ to $0.5$ or $2$ instead raises energy
and strain above the baseline. The largest deterioration occurs at
$\kappa=2$, showing that the physical-quality gains are sensitive to
the path's time-scale parameter under the fixed sampling protocol.

\end{document}